%% file: document.tex
\documentclass[journal]{IEEEtran}
\usepackage{subfiles}
\usepackage{times}
\usepackage{soul}
\usepackage{url}
\usepackage[utf8]{inputenc}
\usepackage{graphicx}
\usepackage{amsmath}
\usepackage{amsthm}
\usepackage{amssymb}
\usepackage{mathtools}
\usepackage{booktabs}
\usepackage{algorithm}
\usepackage{algorithmic}
\usepackage[switch]{lineno}
\usepackage{lettrine}
\usepackage{multirow}
\usepackage{longtable}
\usepackage{array}
\usepackage{xcolor}
\usepackage[caption=false,font=footnotesize]{subfig}

\usepackage[breaklinks, colorlinks, 
citecolor=blue,
linkcolor=red,
urlcolor=blue]{hyperref}

\newcommand{\TV}{\mathrm{TV}}
\newcommand{\HV}{\operatorname{HV}}

\begin{document}

\subfile{APBMM}

\subfile{supplementary2}

\end{document}

%% file: APBMM.tex
\title{AP-BMM: Approximating Capability-Cost Pareto Sets of LLMs via Asynchronous Prior-Guided Bayesian Model Merging}
	\author{Kesheng~Chen, Yamin~Hu, Zhenqian~Zhu, Yiya~Diao, Wenjian~Luo,~\IEEEmembership{Senior Member,~IEEE}%
	\thanks{Kesheng~Chen, Yamin~Hu, Zhenqian~Zhu, Yiya~Diao and Wenjian~Luo are with Guangdong Provincial Key Laboratory of Novel Security Intelligence Technologies, Institute of Cyberspace Security, School of Computer Science and Technology, Harbin Institute of Technology, Shenzhen 518055, China. (e-mail: 22s151138@stu.hit.edu.cn, huyamin@hit.edu.cn, 23b351010@stu.hit.edu.cn, diaoyiyacug@gmail.com, luowenjian@hit.edu.cn).}
}
\maketitle
\begin{abstract}
	LLM services need to offer a family of models spanning different capability--cost trade-offs to accommodate diverse user preferences. Model merging offers a practical way to construct such a model family by combining a reasoning-enhanced model with an instruction-tuned model. Compared with model-level merging, layer-wise merging offers finer control over the capability--cost trade-off by assigning different merge weights to individual layers. However, it introduces two practical challenges: the layer-wise search space is high-dimensional, and existing methods rarely exploit informative signals from the source models; moreover, the highly variable runtime of LLM evaluations makes synchronous batch optimization inefficient by leaving GPU resources idle while waiting for slow evaluations. To address these challenges, we propose Asynchronous Prior-Guided Bayesian Model Merging (AP-BMM), which formulates layer-wise merging as a multi-objective optimization problem to approximate a Pareto set of merged LLMs, yielding a family of Pareto-optimal merged models with diverse capability--cost trade-offs. AP-BMM leverages parameter and activation discrepancies between the source models to guide the early layer-wise search, employs asynchronous pending-aware Bayesian optimization to maximize GPU utilization through asynchronous evaluations and to select high-quality candidates via pending awareness, and applies lightweight ranking over an oversized candidate pool to improve Pareto-front coverage. Under fixed evaluation budgets, AP-BMM achieves higher hypervolume (HV) and broader Pareto-front coverage than synchronous layer-wise and representative model-level merging methods, while reducing wall-clock time through higher GPU utilization.	

	Code is available at the following anonymous repository: \url{https://anonymous.4open.science/r/AP-BMM}.
	
\end{abstract}
\begin{IEEEkeywords}
	Large Language Models, Model Merging, Multi-objective Optimization, Bayesian Optimization
\end{IEEEkeywords}
	
\section{Introduction}
\label{sec:introduction}
	
\lettrine{L}{arge} language models (LLMs) have substantially advanced natural language processing, with chain-of-thought (CoT) prompting serving as a key technique for eliciting more structured and effective reasoning~\cite{wei2022chain,wangself,zhouleast,ye2026re}. However, stronger reasoning often comes with longer generated traces, which increase generated-token cost and inference latency~\cite{sui2025stop}. In practice, LLM users may have different requirements for reasoning capability and generated-token cost, motivating LLM services to support a range of model variants rather than a single fixed model. For instance, a user may prefer a reasoning-enhanced model that produces longer but more accurate responses, while another user may prioritize a lower-cost instruction-tuned model that generates shorter responses~\cite{yang2025qwen3}.
	
Against this background, model merging has emerged as a practical way to construct a model family that covers different \emph{capability--cost} preferences by combining a reasoning-enhanced model with an instruction-tuned model~\cite{wu2025revisiting,team2025kimi}. This naturally motivates formulating capability--cost model merging as a Pareto-set approximation problem, where each Pareto-optimal solution corresponds to a merged model with a distinct balance between reasoning capability and generated-token cost.
	
Existing model merging methods~\cite{utans1996weight,ilharco2022editing,yadav2023ties,yu2024language,deep2024della} predominantly rely on coarse model-level merging operators, where a single global merging weight is applied uniformly across all layers. Such a uniform treatment overlooks the heterogeneous roles of different layers in reasoning and instruction following. Consequently, model-level merging often produces only a limited range of capability--cost trade-offs, leaving intermediate regions of the capability--cost spectrum underexplored.

\begin{figure}[!t]
	\centering
	\includegraphics[width=1.0\columnwidth]{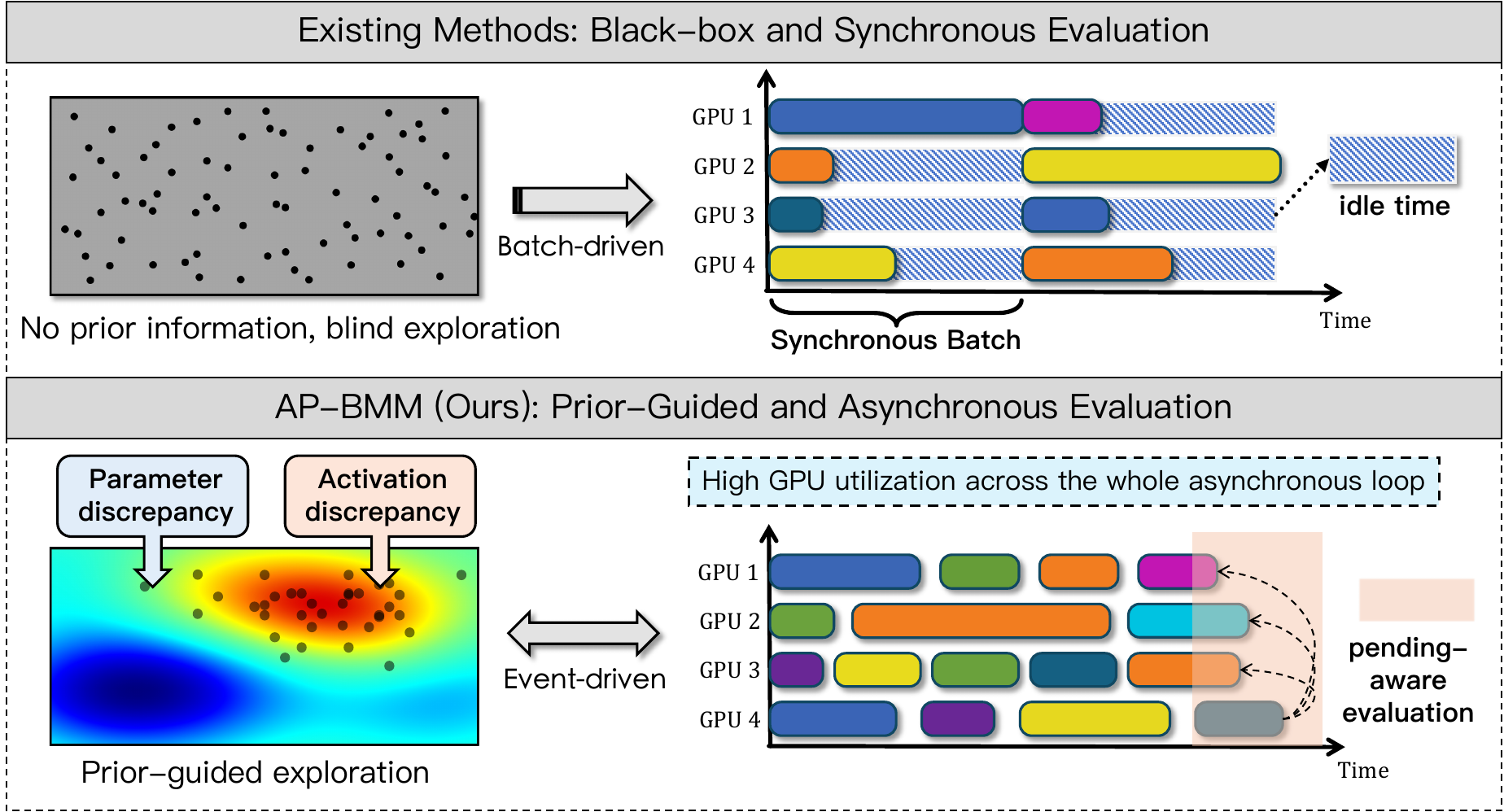}
	\caption{Comparison between existing black-box synchronous merging methods and our prior-guided asynchronous merging method.}
	\label{fig:motivation}
\end{figure}

Recent layer-wise model merging methods partially address this limitation by searching for per-layer merging weights using evolutionary computation or Bayesian optimization~\cite{akiba2025evolutionary,li2025s}. Compared with model-level merging, they provide finer control over layer-wise interpolation and enable a richer family of capability--cost trade-offs. However, under tight evaluation budgets, they still suffer from two key limitations: (1) Lack of Prior Guidance: Existing methods fail to exploit informative signals already present in source models, treating all layers as equally unknown at initialization. Consequently, the optimizer must discover from scratch which layers are most influential for the capability--cost trade-off, leading to increasingly inefficient search as model depth grows. (2) Synchronous Evaluation Bottleneck: Current automated merging frameworks rely on synchronous batch optimization, such as CMA-ES~\cite{hansen2003reducing} or batch $q$NEHVI~\cite{daulton2020differentiable,daulton2022lognehvi}. Since evaluation time is dominated by autoregressive token generation, merged candidates with different reasoning depths exhibit highly variable runtimes, leaving GPU workers idle while waiting for the slowest evaluation in each batch. These limitations are illustrated in Fig.~\ref{fig:motivation}.

To address these limitations, we propose \textbf{A}synchronous \textbf{P}rior-guided \textbf{B}ayesian \textbf{M}odel \textbf{M}erging (AP-BMM), a multi-objective optimization framework for approximating the capability--cost Pareto set of LLMs via layer-wise model merging, as illustrated in Fig.~\ref{fig:motivation}. AP-BMM leverages parameter and activation discrepancies between the source models to guide early layer-wise search toward more promising layers. It further employs asynchronous pending-aware Bayesian optimization to maximize GPU utilization through asynchronous evaluations while selecting high-quality candidates by accounting for pending evaluations. Moreover, AP-BMM applies lightweight ranking over an oversized candidate pool to improve capability--cost Pareto-front coverage.

Our contributions are as follows:
\begin{itemize}
\item We formulate capability--cost LLM model merging as a layer-wise multi-objective optimization problem, and introduce informative priors derived from parameter and activation discrepancies to guide early exploration toward layers that are more likely to influence the capability--cost trade-off.
\item We propose an asynchronous pending-aware Bayesian optimization framework that maximizes GPU utilization through asynchronous evaluations, selects high-quality candidates by accounting for pending evaluations, and further improves Pareto-front coverage via lightweight ranking over an oversized candidate pool.
\item We conduct comprehensive experiments against representative model-level and layer-wise merging methods, together with targeted ablation studies. AP-BMM achieves higher hypervolume (HV), broader Pareto-front coverage, and shorter wall-clock time under fixed evaluation budgets, while recovering trade-off regions that remain unattainable by model-level merging methods even with doubled evaluation budgets.
\end{itemize}

The rest of this paper is organized as follows. Section~\ref{sec:background} reviews model merging and multi-objective optimization. Section~\ref{sec:method} presents the AP-BMM framework, including the problem formulation, prior-guided GP surrogate, and asynchronous pending-aware Bayesian optimization. Section~\ref{sec:experimental_setup} and Section~\ref{sec:main_results} describe the experimental setup and results. Finally, Section~\ref{sec:conclusion} concludes the paper and discusses future directions.

\section{Background}
\label{sec:background}
\subsection{Model Merging}

Model merging aims to combine the parameters of multiple neural network models to integrate their complementary capabilities without retraining from scratch~\cite{yang2024model}. Given a set of source models derived from a common base model, the objective is to directly construct a merged model by merging the parameters of the source models.

Existing model merging methods~\cite{utans1996weight,ilharco2022editing,yadav2023ties,yu2024language,deep2024della} primarily operate at the model level. Weight averaging~\cite{utans1996weight} constructs a merged model by taking a weighted average of corresponding parameters. Task arithmetic~\cite{ilharco2022editing} extracts a task vector by subtracting the base-model parameters from a fine-tuned model and then adds a scaled version of this vector back to the base model. More advanced methods, including TIES-Merging~\cite{yadav2023ties}, DARE~\cite{yu2024language}, and DELLA-Merging~\cite{deep2024della}, further incorporate sparsification, sign agreement, and pruning mechanisms to mitigate parameter conflicts and redundancy. Although computationally efficient, these methods operate at a coarse model-level granularity and rely on handcrafted hyperparameters, limiting their ability to systematically explore the capability--cost trade-off.

Recent methods move beyond fixed global merging rules by optimizing layer-wise merging weights. Akiba et al.~\cite{akiba2025evolutionary} employ CMA-ES~\cite{hansen2003reducing} to optimize these merging weights; we refer to this method as TIES-DARE in this paper. However, it combines multiple objectives into a single scalar objective, resulting in a single-objective optimization procedure. Although different merging weights can produce different trade-offs, the method does not directly approximate the capability--cost Pareto set. In contrast, MOMM~\cite{li2025s} formulates model merging as a multi-objective optimization problem and employs Bayesian optimization with the Log-Noisy Expected Hypervolume Improvement ($q$LogNEHVI) acquisition function~\cite{daulton2022lognehvi} to directly identify Pareto-optimal merged models across conflicting objectives.

Despite these advances, both TIES-DARE~\cite{akiba2025evolutionary} and MOMM~\cite{li2025s} rely on synchronous optimization loops: They dispatch a batch of candidate models and wait for all evaluations to finish before updating the optimizer. In LLM benchmarking, evaluation latency is largely determined by autoregressive token generation. Because layer-wise merging can produce both reasoning-heavy models with long decoding trajectories and lower-cost models with much shorter outputs, evaluation times may vary substantially across candidates. This synchronization leaves GPU workers idle while waiting for the slowest evaluation to complete, reducing the number of candidates that can be evaluated within a fixed time budget. Furthermore, these methods treat the high-dimensional layer-wise search space largely as a black box, without exploiting informative prior knowledge already available from the source models to guide early search.

\subsection{Multi-objective Optimization}
Multi-objective optimization aims to find a set of solutions that represent the optimal trade-offs among conflicting objectives~\cite{deb2002fast,chen2024evolutionary}. Formally, it optimizes a vector-valued objective function
$F(\mathbf{x}) = (f_1(\mathbf{x}), f_2(\mathbf{x}), \ldots, f_K(\mathbf{x}))$
over a feasible decision space. The key concepts are as follows:
\begin{itemize}
    \item \textbf{Pareto dominance:} $\mathbf{x}$ dominates $\mathbf{y}$ if $f_i(\mathbf{x}) \leq f_i(\mathbf{y})$ for all $i$ and $f_j(\mathbf{x}) < f_j(\mathbf{y})$ for at least one $j$ under minimization; the inequalities are reversed under maximization.
    \item \textbf{Pareto optimality:} A solution is Pareto optimal if no other feasible solution dominates it.
    \item \textbf{Pareto set (PS):} The set of all Pareto-optimal solutions in the decision space.
    \item \textbf{Pareto front (PF):} The image of the Pareto set in the objective space, representing the optimal trade-offs among objectives.
\end{itemize}	

Multi-objective optimization methods, including evolutionary algorithms such as MOEA/D~\cite{zhang2007moea} and Bayesian optimization methods such as batch $q$NEHVI~\cite{daulton2022lognehvi}, typically evaluate candidate solutions in synchronous batches. However, this synchronization becomes inefficient when evaluation times vary substantially across candidates, as faster evaluations must wait for the slowest one before the optimizer can proceed. Asynchronous multi-objective optimization is therefore particularly well suited to problems with expensive and heterogeneous evaluation costs, such as LLM evaluations.

\section{AP-BMM: Asynchronous Prior-Guided Bayesian Model Merging}
\label{sec:method}

In this section, we present AP-BMM. We first provide an overview of the framework and formulate the layer-wise capability--cost model merging problem. We then describe its two key components: the prior-guided GP surrogates and the pending-aware asynchronous Bayesian optimization.

\subsection{Overview}
AP-BMM aims to approximate the capability--cost Pareto set of LLMs through model merging, i.e., to construct a collection of non-dominated merged models that span diverse trade-offs between reasoning capability and generated-token cost. AP-BMM takes as input a shared base model and two derived models representing opposite ends of the capability--cost spectrum: a reasoning-enhanced model with stronger reasoning capability but higher generated-token cost, and an instruction-tuned model with lower generated-token cost but comparatively weaker reasoning capability.  Given these inputs, AP-BMM formulates layer-wise model merging as a multi-objective optimization problem, where each candidate is a layer-wise merge-weight vector and the objectives measure the reasoning capability and generated-token cost of the resulting merged model. Notably, although this paper focuses on merging two source models, the formulation can naturally extend to more source models.

Directly evaluating many candidate merge-weight vectors is expensive because each evaluation requires running the merged model on a benchmark suite. AP-BMM therefore uses Bayesian optimization to reduce the number of expensive evaluations.
Specifically, a Gaussian process (GP) surrogate serves as a probabilistic proxy by modeling the mapping from layer-wise merge-weight vectors to the corresponding capability--cost objectives. Based on the surrogate posterior, an acquisition function is optimized to identify the most promising candidate for evaluation. Once the evaluation is completed, the resulting observations are incorporated into the completed evaluation set, and the GP surrogate is refit to guide the next round of candidate selection.

To make Bayesian optimization more efficient, AP-BMM addresses two coupled bottlenecks: search inefficiency and system inefficiency. Search inefficiency arises from the high-dimensional decision space, where each layer has its own merging weights. Treating all layers as equally uninformative ignores useful discrepancy signals already present in the source models. AP-BMM therefore inspects the source models before expensive benchmark evaluations, computes parameter- and activation-based discrepancies, and converts them into an importance prior over layers. This prior initializes the GP lengthscales, making the early search more sensitive to layers that are more likely to affect the capability--cost trade-off.

System inefficiency arises from heterogeneous evaluation times across candidate merged models.
Reasoning-enhanced candidates typically generate longer reasoning traces and therefore require substantially longer evaluation times, whereas lower-cost candidates often finish much earlier.
AP-BMM therefore executes the search asynchronously with a continuously running GPU worker pool. Instead of waiting for all candidates in a batch to finish, AP-BMM updates the completed evaluation set and refits the GP whenever any worker returns an evaluation result. The newly idle worker is then assigned another candidate selected by maximizing a pending-aware acquisition function, which accounts for candidates that are already under evaluation. This event-driven loop keeps GPUs active and lets the search incorporate new evidence as soon as it becomes available.

Algorithm~\ref{alg:async-bo} summarizes the AP-BMM procedure. First, the GPU worker pool is initialized, and the layer-wise importance prior is computed from the source models. Then, an initial set of layer-wise merge-weight vectors is evaluated to form the initial completed evaluation set. The main loop continues until the evaluation budget is exhausted: if there are idle workers and the total number of completed and pending evaluations is below the budget, new candidates are generated by optimizing a pending-aware acquisition function, ranked using the proposed gap reward and proximity penalty, and dispatched to idle workers; whenever a worker returns an evaluation result, the completed evaluation set is updated. Finally, the approximated capability--cost Pareto set is constructed from the completed evaluations.

\begin{algorithm}[htbp]
    \caption{AP-BMM: Asynchronous Prior-Guided Bayesian Model Merging}
    \label{alg:async-bo}
    \begin{algorithmic}[1]
        \REQUIRE Base model $\mathcal{M}_{\mathrm{base}}$; reasoning-enhanced model $\mathcal{M}_{\mathrm{cap}}$; instruction-tuned model $\mathcal{M}_{\mathrm{cost}}$; evaluation budget $B$; GPU worker pool $W$; benchmark suite $\mathcal{S}$
        \ENSURE Approximation of the capability--cost Pareto set of merged LLMs $\mathcal{F}_{\mathrm{final}}$
        \STATE Initialize the GPU worker pool $W$: keep a warm vLLM server for each worker to hot-load newly merged model weights.
        \STATE Compute layer-wise parameter and activation discrepancies from $\mathcal{M}_{\mathrm{base}},\mathcal{M}_{\mathrm{cap}},\mathcal{M}_{\mathrm{cost}}$, and construct the layer-wise importance prior $\{s_l\}$ from these discrepancies (Section~\ref{sec:prior}).
		\STATE Evaluate an initial set of layer-wise merge-weight vectors $\{\mathbf{x}_i\}$ on the benchmark suite $\mathcal{S}$ to obtain the corresponding objective vectors $\{\mathbf{y}_i\}$.
		\STATE Initialize the completed set as $\mathcal{D}\leftarrow\{(\mathbf{x}_i,\mathbf{y}_i)\}$ and the pending set as $\mathcal{P}\leftarrow\emptyset$.
        \WHILE{$|\mathcal{D}| < B$}
            \IF{$|\mathcal{D}| + |\mathcal{P}| < B$ \textbf{and} at least one worker in $W$ is idle}
                \STATE Fit GP surrogates on $\mathcal{D}$ with lengthscales initialized by the layer-wise importance prior $\{s_l\}$ (Section~\ref{sec:gp}).
                \STATE Generate an oversized candidate merge-weight vector pool $\mathcal{C}$ by optimizing $\alpha_{\mathrm{async}}(X\mid\mathcal{D},\mathcal{P})$; rank candidates using our proposed gap reward and proximity penalty; select high-ranked candidates to form the dispatch set $X^\star$ (Section~\ref{sec:async_opt}).
                \STATE Add $X^\star$ to $\mathcal{P}$; for each $\mathbf{x}\in X^\star$, construct the corresponding merged model and dispatch it to an idle worker.
                \STATE Each worker evaluates its assigned merged model on benchmark suite $\mathcal{S}$ and returns the objective vector asynchronously.
            \ENDIF
            \IF{worker $w$ returns observation $\mathbf{y}$ for merge-weight vector $\mathbf{x}$}
                \STATE $\mathcal{D}\leftarrow\mathcal{D}\cup\{(\mathbf{x},\mathbf{y})\}$; $\mathcal{P}\leftarrow\mathcal{P}\setminus\{\mathbf{x}\}$.
            \ENDIF
        \ENDWHILE
        \STATE Construct the Pareto set $\mathcal{F}_{\mathrm{final}}$ from $\mathcal{D}$.
        \RETURN $\mathcal{F}_{\mathrm{final}}$
    \end{algorithmic}
\end{algorithm}

\subsection{Problem Formulation}
\label{sec:problem_formulation}

Given a reasoning-enhanced model $\mathcal{M}_{\mathrm{cap}}$ and an instruction-tuned model $\mathcal{M}_{\mathrm{cost}}$, both derived from the same base model $\mathcal{M}_{\mathrm{base}}$, we construct a merged model by combining their layer-wise parameter updates. For each layer, the merge weights assigned to the two source models are constrained to sum to one. Therefore, only one independent merge weight is required per layer. We parameterize the merged model by a layer-wise merge-weight vector $\mathbf{x}=(x_1,\ldots,x_L)\in[0,1]^L$, where $L$ is the number of Transformer layers and $x_l$ denotes the merge weight assigned to the reasoning-enhanced model at layer $l$. Accordingly, the merge weight assigned to the instruction-tuned model is $1-x_l$. 

The layer-wise merging operation is defined as
\begin{equation}
\mathcal{M}_{\mathrm{merge},l}
=
\mathcal{M}_{\mathrm{base},l}
+
x_l\TV_{\mathrm{cap},l}
+
(1-x_l)\TV_{\mathrm{cost},l},
\label{eq:layerwise_merge_param}
\end{equation}
where $\TV_{\mathrm{cap},l}$ and $\TV_{\mathrm{cost},l}$ denote the task vectors of the reasoning-enhanced and instruction-tuned models at layer $l$, respectively, defined as
\begin{equation}
\begin{aligned}
\TV_{\mathrm{cap},l}
&=
\mathcal{M}_{\mathrm{cap},l}
-
\mathcal{M}_{\mathrm{base},l},\\
\TV_{\mathrm{cost},l}
&=
\mathcal{M}_{\mathrm{cost},l}
-
\mathcal{M}_{\mathrm{base},l}.
\end{aligned}
\label{eq:task_vector}
\end{equation}
Substituting Eq.~\eqref{eq:task_vector} into Eq.~\eqref{eq:layerwise_merge_param} yields the equivalent convex-combination form
\begin{equation}
\mathcal{M}_{\mathrm{merge},l}
=
x_l\mathcal{M}_{\mathrm{cap},l}
+
(1-x_l)\mathcal{M}_{\mathrm{cost},l}.
\end{equation}

We formulate layer-wise model merging as a two-objective maximization problem:
\begin{equation}
    \max_{\mathbf{x}\in\mathcal{X}}
    F\!\left(\mathcal{M}_{\mathrm{merge}}(\mathbf{x})\right)
    =
    \left(
    f_{\mathrm{cap}}\!\left(\mathcal{M}_{\mathrm{merge}}(\mathbf{x})\right),
    f_{\mathrm{cost}}\!\left(\mathcal{M}_{\mathrm{merge}}(\mathbf{x})\right)
    \right),
\end{equation}
where $\mathbf{x}$ is the layer-wise merge-weight vector, $\mathcal{X}$ denotes the decision space, $\mathcal{M}_{\mathrm{merge}}(\mathbf{x})$ is the merged model induced by $\mathbf{x}$, $f_{\mathrm{cap}}(\cdot)$ measures reasoning capability, and $f_{\mathrm{cost}}(\cdot)$ measures generated-token cost.

Each merged model is evaluated using a set of metrics for each objective. For the capability objective, these metrics include predictive accuracies on the selected benchmarks. For the cost objective, they include the average number of generated tokens per instance across the same benchmarks.

Since these metrics often have different numerical ranges, we normalize all objectives to a common scale.
For each objective, we choose an expert reference and a vanilla reference, and express the merged model's score as its position on the scale defined by these two references.
Let $\mathcal{M}^{(k)}_{\mathrm{expert}}$ be the expert model specialized for the $k$-th objective, and $\mathcal{M}^{(k)}_{\mathrm{vanilla}}$ be the corresponding vanilla model. The reference-scaled objective is
	\begin{equation}
		\begin{aligned}
			f_{k}(\mathcal{M}_{\mathrm{merge}})
			&=
			\frac{1}{N^{(k)}}
			\sum_{s^{(k)} \in S^{(k)}}
			\frac{
				s^{(k)}(\mathcal{M}_{\mathrm{merge}}) - s^{(k)}(\mathcal{M}^{(k)}_{\mathrm{vanilla}})
			}{
				s^{(k)}(\mathcal{M}^{(k)}_{\mathrm{expert}}) - s^{(k)}(\mathcal{M}^{(k)}_{\mathrm{vanilla}})
			},
		\end{aligned}
		\label{eq:reference_scaled_fk}
	\end{equation}
	where $S^{(k)}$ is the metric set associated with the $k$-th objective, with each benchmark score treated as a separate metric when multiple benchmarks are used; $s^{(k)}(\cdot)$ is one scalar metric in that set; and $N^{(k)}$ is the number of metrics for the $k$-th objective.
	The reference-scaled objective maps the vanilla reference to $0$ and the expert reference to $1$, so that improvements across objectives are compared on a common scale. 

	For each objective, we define an expert reference model and a vanilla reference model to construct the reference-scaled objective. For the capability objective, the reasoning-enhanced model serves as the expert reference, while the instruction-tuned model serves as the vanilla reference. For the generated-token-cost objective, the roles are reversed: the instruction-tuned model serves as the expert reference, whereas the reasoning-enhanced model serves as the vanilla reference. Notably, this choice transforms the original cost minimization objective into a maximization objective, ensuring that larger values of the reference-scaled objective consistently indicate better performance across all objectives.

\subsection{Prior-guided GP Surrogates}
\label{sec:prior-guided}
In AP-BMM, Bayesian optimization uses GP surrogates to estimate the objective values of candidate layer-wise merge-weight vectors before running expensive LLM evaluations. The efficiency of the GP-guided search depends on how quickly the GP surrogate identifies which layers matter. Layer-wise merge weights are unlikely to affect reasoning capability and generated-token cost equally: changing the weight of one layer may strongly affect reasoning capability, while changing another may mainly influence generated-token cost or have little observable effect. To guide optimization more efficiently, we extract parameter and activation discrepancy signals from the input source models and convert them into an importance prior over layers. This prior is used to initialize the GP surrogates, giving the GP a structured initial sensitivity over layers. As a result, the GP does not need to infer layer relevance entirely from expensive evaluations, which can accelerate the early search. Importantly, the prior does not exclude any feasible merge weights; it only initializes the surrogate to be more sensitive to informative layers at the beginning of the search. As more evaluations are collected, the GP can revise this initial bias based on observed capability--cost results.

\subsubsection{Discrepancy-based importance prior} 
\label{sec:prior}

We first describe how to compute the parameter and activation discrepancies, and then show how they are combined into a layer-wise importance prior.

\paragraph{Parameter discrepancy} 
Given a reasoning capability model and a instruction-tuned model derived from the same base model, we compare their layer-wise task vectors to measure how differently they deviate from the shared base model. This comparison is useful because layers that differ more strongly between the two source models are more likely to encode behavior relevant to the capability--cost tradeoffs.

\begin{figure}[!t]
\centering
\subfloat[Parameter discrepancy\label{fig:para_prior}]{%
\begin{minipage}[]{\columnwidth}
	\centering
	\includegraphics[width=\linewidth]{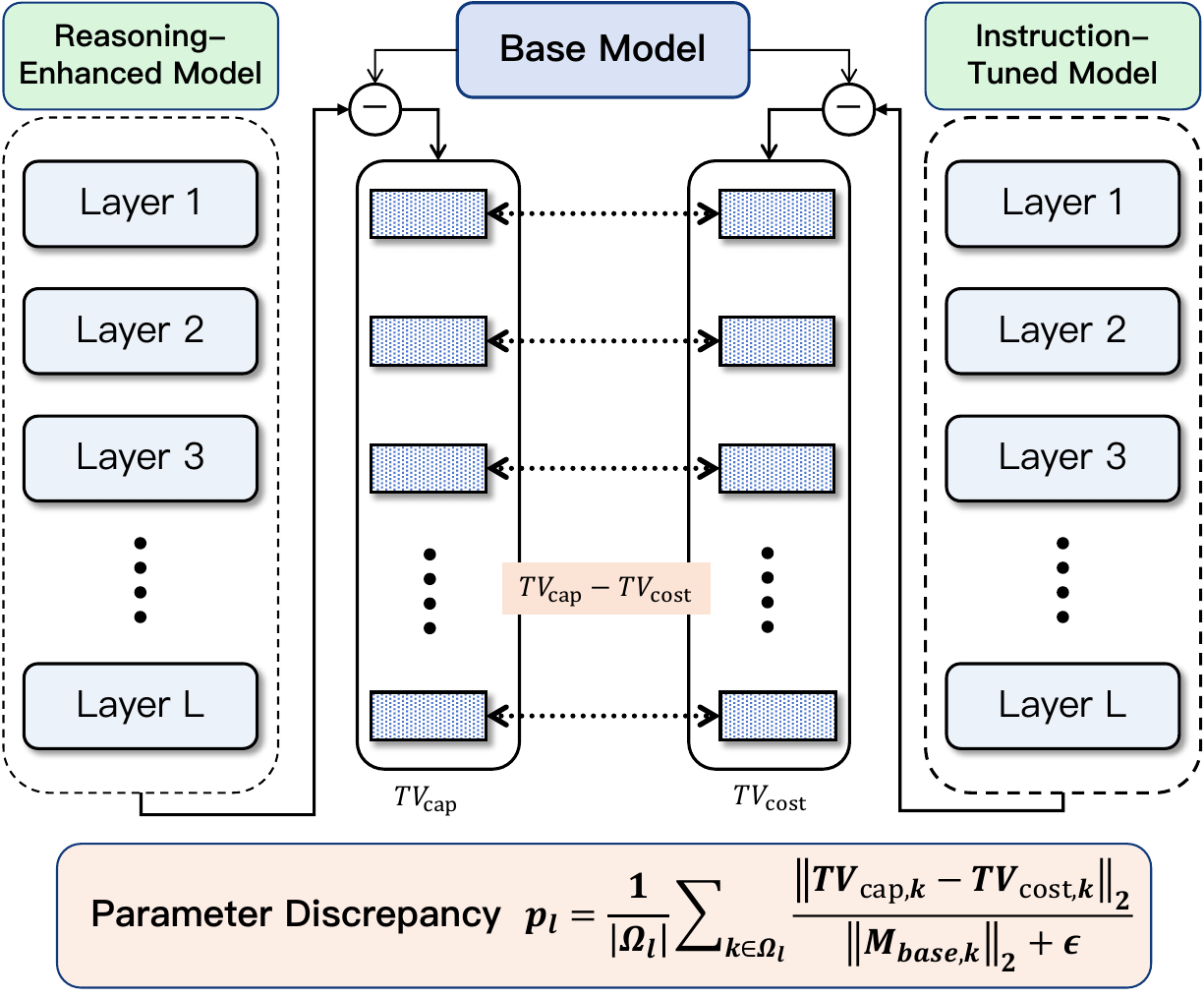}%
\end{minipage}
}
\hfill
\subfloat[Activation discrepancy\label{fig:act_prior}]{%
\begin{minipage}[]{\columnwidth}
			\centering
			\includegraphics[width=\linewidth]{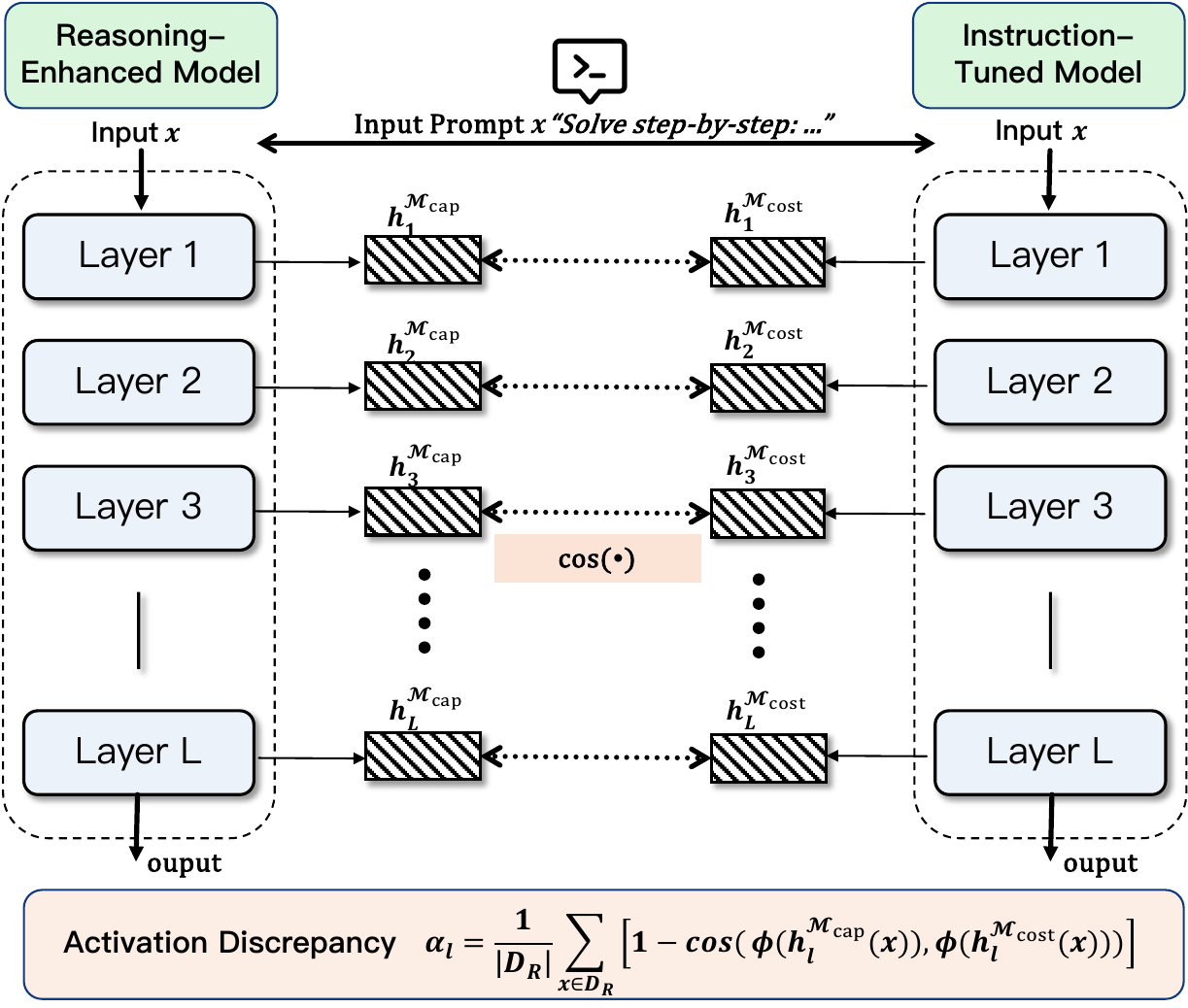}%
		\end{minipage}
}
\caption{Parameter and activation discrepancies.}
\label{fig:prior_components}
\end{figure}

The parameter discrepancy score of layer $l$ is defined as
\begin{equation}
    p_l
    =
    \frac{1}{|\Omega_l|}
    \sum_{k \in \Omega_l}
    \frac{\|\TV_{\mathrm{cap},k} - \TV_{\mathrm{cost},k}\|_2}{\|\mathcal{M}_{\text{base},k}\|_2 + \varepsilon},
\end{equation}
where $\Omega_l$ is the index set of parameter tensors assigned to layer $l$, $k$ indexes a tensor in this set, $\TV_{\mathrm{cap},k}$, $\TV_{\mathrm{cost},k}$ are the task-vector tensors of the reasoning-enhanced model and the instruction-tuned model, respectively, both computed with respect to the shared base-model tensor $\mathcal{M}_{\text{base},k}$, $\|\cdot\|_2$ denotes the Euclidean norm, and $\varepsilon$ is a small constant for numerical stability. Fig.~\ref{fig:para_prior} illustrates the computation of the parameter-based prior.

To keep the discrepancy scores numerically stable and comparable across layers, we rescale them as
\begin{equation}
    s_l^{(p)} = \mathcal{N}(p_l),
\end{equation}
where $\mathcal{N}(\cdot)$ denotes a robust rescaling operator based on median centering, median-absolute-deviation scaling, and sigmoid squashing, following standard robust-statistics practice~\cite{huber2009robust}. This transformation reduces the influence of unusually large layer discrepancies and maps the discrepancy scores to a stable range.

\paragraph{Activation discrepancy}
Parameter discrepancy alone may not fully capture functional differences between the reasoning-enhanced and instruction-tuned models. A small parameter change can induce substantial shifts in hidden representations, whereas a large parameter change may have only a limited effect on the model's intermediate computation. We therefore also measure activation discrepancy between the two source models. By comparing their hidden activations on the same prompt set, we identify layers whose representations differ more strongly and may therefore require more careful search.

The activation discrepancy of layer $l$ is defined as
\begin{equation}
    \alpha_{l}
    =
    \frac{1}{|D_R|}
    \sum_{x \in D_R}
    \left[
    1 - \cos\!\Big(\phi(h_l^{\mathcal{M}_{\mathrm{cap}}}(x)), \phi(h_l^{\mathcal{M}_{\mathrm{cost}}}(x))\Big)
    \right],
\end{equation}
where $D_R$ is a reasoning-oriented prompt set, $x$ is a prompt in $D_R$, $h_l^{\mathcal{M}_{\mathrm{cap}}}(x)$ and $h_l^{\mathcal{M}_{\mathrm{cost}}}(x)$ are the hidden representations at layer $l$ produced by the reasoning-enhanced and instruction-tuned models, respectively, and $\phi(\cdot)$ is a pooling operator. Fig.~\ref{fig:act_prior} illustrates the computation of the activation discrepancy. For efficient computation, the average over $D_R$ can be estimated in mini-batches: we compute the discrepancy for each batch and then average the batch-level scores to obtain $\alpha_l$. We apply the same robust rescaling operator used for the parameter discrepancy to the activation discrepancy:
\begin{equation}
    s_{l}^{(a)} = \mathcal{N}(\alpha_{l}).
\end{equation}

\paragraph{Fused importance prior}
We fuse the parameter and activation discrepancies into a layer-wise importance prior using a weighted sum:
\begin{equation}
    c_l = k_1 s_l^{(p)} + k_2 s_l^{(a)},
\end{equation}
where $c_l$ denotes the fused importance prior of layer $l$, and $k_1,k_2\ge0$ are normalized weighting coefficients satisfying $k_1+k_2=1$, which control the relative contributions of the parameter and activation discrepancies.

Because neighboring layers often exhibit correlated behavior rather than acting independently, we further smooth the layer-wise importance prior by aggregating information from adjacent layers:
\begin{equation}
    \mathcal{S}(c_l)=
    \begin{cases}
        c_l, & l=1 \text{ or } l=L,\\
        0.25\,c_{l-1}+0.5\,c_l+0.25\,c_{l+1}, & 1<l<L.
    \end{cases}
\end{equation}
where $\mathcal{S}(\cdot)$ denotes a local smoothing operator, while the boundary layers remain unchanged. The resulting smoothed layer-wise importance prior is then given by
\begin{equation}
s_l=\mathcal{S}(c_l).
\end{equation}

\subsubsection{GP Surrogate Training and Prediction}
\label{sec:gp}

We first describe how the resulting layer-wise importance prior is used to initialize the GP surrogates, and then explain how the GP surrogates are trained and used to make predictions.

\paragraph{GP Surrogate Initialization}

The GP surrogate uses a kernel function to measure similarity between two merge-weight vectors, whose lengthscales determine the surrogate's sensitivity to changes along each layer dimension. Unlike a standard isotropic kernel~\cite{Rasmussen2006gaussian}, which shares a single lengthscale across all input dimensions, AP-BMM adopts automatic relevance determination (ARD)~\cite{Rasmussen2006gaussian} to assign an independent lengthscale to each layer. Since only a limited number of evaluations are available during the early stages of Bayesian optimization, the initialization of these lengthscales can strongly influence the search behavior. AP-BMM initializes the ARD lengthscales using the fused layer-wise importance prior, so that layers with higher prior importance are regarded as more influential at the beginning of the search. As additional evaluations become available, all lengthscales remain learnable and are updated during GP training.

In the GP surrogates in AP-BMM, we use a Mat\'ern-$5/2$ kernel~\cite{Rasmussen2006gaussian,eriksson2021high}. This kernel is commonly used in Bayesian optimization because it provides a moderate smoothness assumption for black-box objective functions~\cite{Rasmussen2006gaussian,eriksson2021high}. For objective $k$, the kernel function for objective $k$ is defined as
\begin{equation}
	\kappa_k(\mathbf{x},\mathbf{x}')
	=
	\sigma_{f,k}^2
	\left(
	1+\sqrt{5}r_k+\frac{5}{3}r_k^2
	\right)
	e^{-\sqrt{5}r_k},
\end{equation}
where $\sigma_{f,k}^2$ denotes the output variance of objective $k$.
The distance term is defined as
\begin{equation}
	r_k(\mathbf{x},\mathbf{x}')
	=
	\sqrt{\sum_{l=1}^{L}\frac{(x_l-x_l')^2}{\ell_{k,l}^2}},
\end{equation}
where $\mathbf{x}$ and $\mathbf{x}'$ are two merge-weight vectors, $x_l$ and $x_l'$ are their merge weights at layer $l$, and $\ell_{k,l}$ is the ARD lengthscale of objective $k$ at layer $l$. A smaller $\ell_{k,l}$ makes the kernel distance more sensitive to changes in layer $l$, so the GP varies more rapidly along that dimension. Conversely, a larger $\ell_{k,l}$ makes the GP smoother with respect to that dimension.

To initialize the ARD lengthscales, we first rank the fused layer-wise importance scores in descending order. Let $\mathrm{rank}_{\downarrow}(s_l)\in\{0,\ldots,L-1\}$ denote the descending rank of layer $l$, where rank $0$ corresponds to the highest importance score. The rank is then normalized as
\begin{equation}
    r_l=\frac{\mathrm{rank}_{\downarrow}(s_l)}{\max(L-1,1)},
\end{equation}
We then map this normalized rank to an initial layer-wise lengthscale:
\begin{equation}
    \ell_l=
    \begin{cases}
        0.5, & r_l \le 0.15,\\
        1.0, & 0.15 < r_l \le 0.40,\\
        2.0, & 0.40 < r_l \le 0.70,\\
        5.0, & \text{otherwise.}
    \end{cases}
\end{equation}
This mapping assigns smaller initial lengthscales to higher-prior layers, making the GP initially more sensitive to their merge weights. Finally, for each objective $k$, the ARD lengthscale is initialized by setting
\begin{equation}
    \ell_{k,l}\leftarrow \ell_l.
\end{equation}

\paragraph{GP Surrogate Training}
AP-BMM fits an independent GP surrogate for each objective using the completed evaluation set $\mathcal{D}=\{(\mathbf{x}_i,\mathbf{y}_i)\}_{i=1}^{n}$. For objective $k$, the surrogate models the mapping from a merge-weight vector to its corresponding objective value:
\begin{equation}
g_k(\mathbf{x}) \sim \mathcal{GP}\left(0,\kappa_k(\mathbf{x},\mathbf{x}')\right),
\end{equation}
where $\kappa_k(\mathbf{x},\mathbf{x}')$ is the covariance kernel.

The GP hyperparameters are learned independently for each objective by maximizing the exact marginal log likelihood:
\begin{equation}
    \hat{\Theta}_k
    =
    \arg\max_{\Theta_k}
    \log p\!\left(\{\mathbf{y}_i\}_{i=1}^{n}\mid \{\mathbf{x}_i\}_{i=1}^{n}, \Theta_k\right),
\end{equation}
where $\Theta_k$ includes the kernel output variance, observation-noise variance, and ARD lengthscales. Notably, the prior-guided lengthscales serve only as the initialization; all GP hyperparameters are jointly optimized during training.

\paragraph{GP Surrogate Prediction}
Once the GP is fitted, it provides predictive distributions over the objective values of unevaluated merge-weight vectors. For any unevaluated merge-weight vector $\mathbf{x}$, the predictive distribution of the random objective value $y_k(\mathbf{x})$ for objective $k$ is $p_k\!\left(y_k(\mathbf{x}) \mid \mathbf{x}\right)$. 

For a candidate batch $X=\{\mathbf{x}^{(1)},\dots,\mathbf{x}^{(q)}\}$, the GP defines the joint predictive distribution over the candidate batch:
\begin{equation}
p\!\left(\mathbf{Y}_{X}\mid \mathcal{D}\right)
=
\prod_{k=1}^{K}
p_k\!\left(\mathbf{Y}_{X,k}\right), 
\end{equation}
where $q$ is the batch size, $\mathbf Y_{X,k}
=
\left(
y_k(\mathbf x^{(1)}),
\ldots,
y_k(\mathbf x^{(q)})
\right)^\top$ denotes the random vector of objective values for objective $k$ over batch $X$, and $p_k\!\left(\mathbf{Y}_{X,k}\right)$ is the joint GP predictive distribution.

\subsection{Pending-aware Asynchronous Bayesian Optimization}
\label{sec:async_opt}

\begin{figure*}[!t]
	\centering
	\subfloat[System-level asynchronous execution.\label{fig:async_architecture}]{%
		\begin{minipage}[]{0.35\textwidth}
			\centering
			\includegraphics[width=\linewidth]{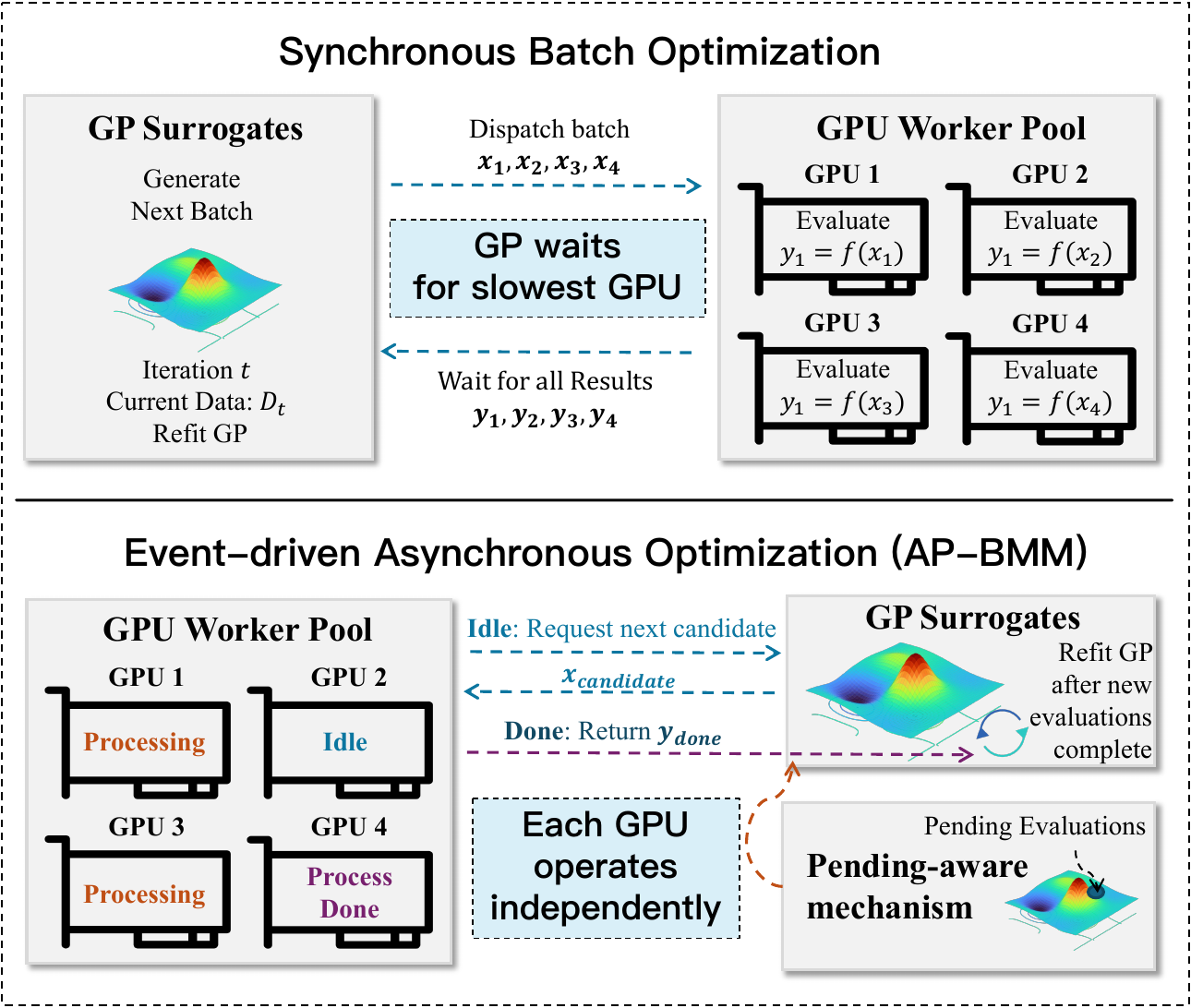}
		\end{minipage}
	}
	\hfill
	\subfloat[Pending-aware candidate selection in AP-BMM. The right panel shows the GPU scheduling, where completed evaluations coexist with in-flight pending evaluations and newly proposed candidates. The left panel maps these evaluations and candidates into the objective space. Ignoring pending evaluations may overestimate the contribution of a new candidate, yielding an inflated hypervolume improvement $\mathrm{HV}_1$. AP-BMM predicts the outcomes of pending candidates with the GP surrogate and includes them in the hypervolume calculation, adjusting the new candidate's additional contribution to $\mathrm{HV}_2$, which better reflects its marginal contribution. \label{fig:async_mechanism}]{	
			\begin{minipage}[]{0.62\textwidth}
			\centering
			\vspace{-52pt}
			\includegraphics[width=\linewidth]{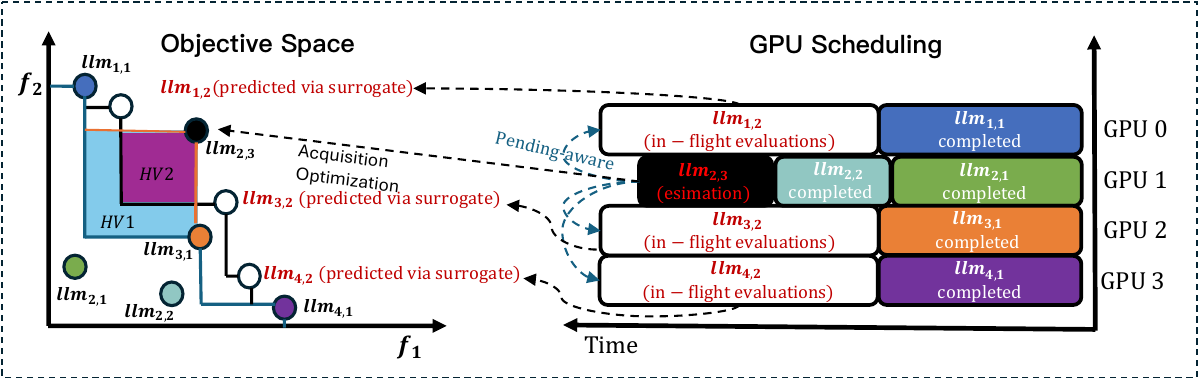}
		\end{minipage}
	}
	\caption{Pending-aware Asynchronous Bayesian Optimization of AP-BMM.}
	\label{fig:async_scheduler_full}
\end{figure*}

Conventional synchronous Bayesian optimization follows a batch-based optimization cycle, as illustrated in the upper panel of Fig.~\ref{fig:async_architecture}. At each iteration, a batch of candidate merge-weight vectors is selected by the GP surrogates and is then dispatched to the GPU worker pool for evaluation. The optimizer waits until all evaluations in the batch have returned before updating the completed evaluation set, refitting the GP surrogates, and generating the next batch. Although this method is simple, it is poorly matched to AP-BMM, where different merged models can have highly variable generation lengths and evaluation times. GPU workers evaluating short-generation candidates may finish early but remain idle until the slowest evaluation completes, delaying both surrogate updates and new candidate generation.

This bottleneck motivates the asynchronous design of AP-BMM, shown in the lower panel of Fig.~\ref{fig:async_architecture}. AP-BMM replaces synchronous batch execution with a continuously running worker pool, where GPUs operate independently without a global synchronization barrier. Each GPU evaluates one candidate at a time, returns its observation as soon as the evaluation finishes, and immediately requests a new candidate when it becomes idle. The surrogate can therefore incorporate completed observations and generate new candidates without waiting for all workers in the previous batch. This design is particularly important for AP-BMM, where generation lengths and evaluation times can vary substantially across merged models.

However, asynchronous evaluation also introduces a new issue: at any time, some candidates may already be under evaluation but have not yet returned objective values. These pending candidates cannot be used as training data for the GP surrogates, but they should still be considered when selecting new candidates, because they may soon occupy regions of the capability--cost trade-off that the optimizer is about to explore again. AP-BMM therefore maintains both a completed evaluation set and a pending evaluation set, and uses a pending-aware acquisition function to score new candidates by their expected additional contribution beyond the regions already being explored by pending evaluations.

\subsubsection{Asynchronous Optimization Loop}
AP-BMM runs an event-driven optimization loop around the GP surrogates. It starts from an initial completed evaluation set, obtained by evaluating a small set of randomly sampled merge-weight vectors, which provides the first training data for the GP surrogates. The optimizer then iteratively updates the GP surrogates as new evaluation results become available and generates new candidates for evaluation until the evaluation budget is exhausted.

At any search step, AP-BMM maintains two evaluation sets: the completed set $\mathcal{D}$, which stores evaluated merge-weight vectors together with their observed objective values, and the pending set $\mathcal{P}$, which stores only the merge-weight vectors that have been dispatched to workers but whose objective values have not yet returned. The GP surrogate is trained only on $\mathcal{D}$. Although excluded from GP training, $\mathcal{P}$ is retained so that the acquisition function can explicitly account for pending evaluations, thereby reducing redundant candidate generation.

The event-driven loop proceeds as follows. When one or more workers become idle, the GP hyperparameters are re-optimized using the current completed evaluation set $\mathcal{D}$. AP-BMM then generates a new batch of candidates by optimizing the pending-aware acquisition function, which accounts for both $\mathcal{D}$ and $\mathcal{P}$. During candidate generation, the acquisition function uses GP predictions to estimate the unknown outcomes of both pending candidates in $\mathcal{P}$ and newly generated candidates. The generated candidates are added to $\mathcal{P}$, and the corresponding merged models are dispatched to idle workers for evaluation.
Whenever a worker finishes an evaluation, AP-BMM appends the returned objective values to the completed evaluation set $\mathcal{D}$ and removes the corresponding merge-weight vector from the pending set $\mathcal{P}$. The final approximation to the capability--cost Pareto set is then obtained by extracting the non-dominated merged models from the completed evaluation set $\mathcal{D}$.

\subsubsection{Scoring Candidate Batches}
AP-BMM scores a candidate batch of merge-weight vectors using the $q$LogNEHVI acquisition function~\cite{daulton2022lognehvi}, which estimates the expected hypervolume improvement over the current capability--cost Pareto front. We extend $q$LogNEHVI to the asynchronous setting by explicitly accounting for pending evaluations.

For a candidate batch $X=\{\mathbf{x}^{(1)},\dots,\mathbf{x}^{(q)}\}$, the acquisition function first extracts the nondominated front from the completed evaluation set: 
\begin{equation} \mathcal{F} = \operatorname{ND}\!\left( \{\mathbf{y}_i \mid (\mathbf{x}_i,\mathbf{y}_i)\in\mathcal{D}\} \right). 
	\label{equation:nd} 
\end{equation}
It then estimates the additional hypervolume contributed by $X$ beyond the region already covered by $\mathcal{F}$ and the regions expected to be occupied by the pending evaluations. The resulting pending-aware acquisition function is 
\begin{equation} \begin{aligned} \alpha_{\mathrm{async}}(X \mid \mathcal{D},\mathcal{P}) &= \mathbb{E}\!\left[ \HV\!\left(\mathcal{F}\cup\mathbf{Y}_{\mathcal{P}}\cup\mathbf{Y}_{X}\right) \right.\\ &\qquad\left. - \HV\!\left(\mathcal{F}\cup\mathbf{Y}_{\mathcal{P}}\right) \right], \end{aligned} 
	\label{eq:async_acquisition} 
\end{equation} 
where $\mathbf{Y}_{\mathcal{P}}$ and $\mathbf{Y}_{X}$ denote posterior samples of the objective values for the pending set $\mathcal{P}$ and the candidate batch $X$, respectively, $\HV(\cdot)$ computes the hypervolume of a set of objective vectors, and the expectation is approximated by Monte Carlo averaging~\cite{balandat2020botorch} over posterior samples of $\mathbf{Y}_{\mathcal{P}}$ and $\mathbf{Y}_{X}$.

This pending-aware acquisition function helps estimate the hypervolume contribution of new candidates more accurately. Fig.~\ref{fig:async_mechanism} illustrates how accounting for pending evaluations changes the estimated contribution of a new candidate from $\mathrm{HV}_1$ to $\mathrm{HV}_2$. When pending candidates are ignored, the optimizer evaluates the new candidate only against the completed evaluation set $\mathcal{D}$ and may obtain an inflated hypervolume increment $\mathrm{HV}_1$. After incorporating the predicted outcomes of pending candidates in $\mathcal{P}$, the estimated hypervolume contribution becomes $\mathrm{HV}_2$, which better reflects the candidate's additional contribution beyond regions already being explored. In this way, the optimizer can reduce duplicate candidates near candidates that are already under evaluation.

\subsubsection{Generating Candidates}
AP-BMM generates new candidates by maximizing the asynchronous acquisition function in Equation~(\ref{eq:async_acquisition}). In practice, however, optimizing the acquisition function alone may still produce candidates concentrated in a narrow region of the capability--cost trade-off, limiting Pareto-front coverage. To encourage more diverse exploration, AP-BMM adopts a two-stage candidate selection procedure. It first optimizes the acquisition function to construct an oversized candidate pool, and then ranks the candidates to select the final batch for evaluation. This decouples candidate quality from candidate diversity: the acquisition function identifies promising candidates, while the ranking stage promotes broader Pareto-front coverage without modifying the acquisition objective itself.

The oversized candidate pool is obtained by approximately maximizing Equation~(\ref{eq:async_acquisition}) with multi-start box-constrained optimization \cite{daulton2020differentiable,daulton2022multi} over $[0,1]^{L}$. We denote the resulting pool by
\begin{equation}
    \mathcal{C}=\{\mathbf{x}_1,\dots,\mathbf{x}_{N_{\mathrm{pool}}}\},
    \quad
    N_{\mathrm{pool}}=\lceil \kappa q \rceil,
\end{equation}
where $q$ is the number of idle workers in the current dispatch event, capped by the remaining evaluation budget, $N_{\mathrm{pool}}$ is the candidate-pool size, and $\kappa \ge 1$ is the pool multiplier.

To rank the candidates in the oversized pool, AP-BMM defines a lightweight ranking score composed of two terms: a gap reward, which favors candidates in sparsely covered regions of the capability--cost trade-off, and a proximity penalty, which discourages candidates that are close to pending evaluations or candidates already selected for the current dispatch.

\paragraph{Gap Reward}
The gap reward encourages candidates whose predicted objective vectors fill gaps in the current nondominated capability--cost Pareto front. To compute this reward, we first extract the nondominated objective vectors from the completed evaluation set and sort them using the capability objective as the primary key and the cost objective as the secondary key. The sorted objective vectors are denoted by
\begin{equation}
    \mathcal{V}=[\mathbf{f}_1,\ldots,\mathbf{f}_{N_{\mathrm{PF}}}],
    \quad
    \mathbf{f}_g=(f^{\mathrm{cap}}_g,f^{\mathrm{cost}}_g),
\end{equation}
where $N_{\mathrm{PF}}$ is the number of nondominated objective vectors, and $f^{\mathrm{cap}}_g$ and $f^{\mathrm{cost}}_g$ denote the capability and cost objective values corresponding to the $g$-th nondominated merge-weight vector, respectively.

Next, we define gaps between adjacent objective vectors in the sorted list. For each adjacent pair $(\mathbf{f}_g,\mathbf{f}_{g+1})$, $g=1,\ldots,N_{\mathrm{PF}}-1$, we compute its midpoint, raw width, and normalized width:
\begin{equation}
\begin{aligned}
    \mathbf{m}_g &= \frac{1}{2}\left(\mathbf{f}_{g}+\mathbf{f}_{g+1}\right),\\
    d_g &= \|\mathbf{f}_{g+1}-\mathbf{f}_{g}\|_2,\\
    w_g &= \frac{d_g}{\max_h d_h}.
\end{aligned}
\end{equation}
Here, $\mathbf{m}_g$ is the center of the gap, $d_g$ is its Euclidean width of the gap, and $w_g$ is the width normalized by the largest gap. A larger $w_g$ indicates a more sparsely covered interval between adjacent nondominated objective vectors. If $\max_h d_h=0$, we set $w_g=0$ for all gaps.

Finally, for each candidate merge-weight vector, we compute how well its predicted objective vector fills these gaps. For a candidate merge-weight vector $\mathbf{x}_i\in\mathcal{C}$, let $\hat{\mathbf{y}}_i=(\hat{y}^{\mathrm{cap}}_i,\hat{y}^{\mathrm{cost}}_i)$ be its GP-predicted objective vector. For each gap $g$, we measure the proximity between $\hat{\mathbf{y}}_i$ and its midpoint $\mathbf{m}_g$ using a Gaussian kernel, and weight this proximity by the normalized gap width $w_g$. 

\begin{figure}[!t]
	\centering
	\subfloat[Gap distance\label{fig:gap_distance}]{%
		\includegraphics[width=0.36\columnwidth,trim=10 5 25 30, clip]{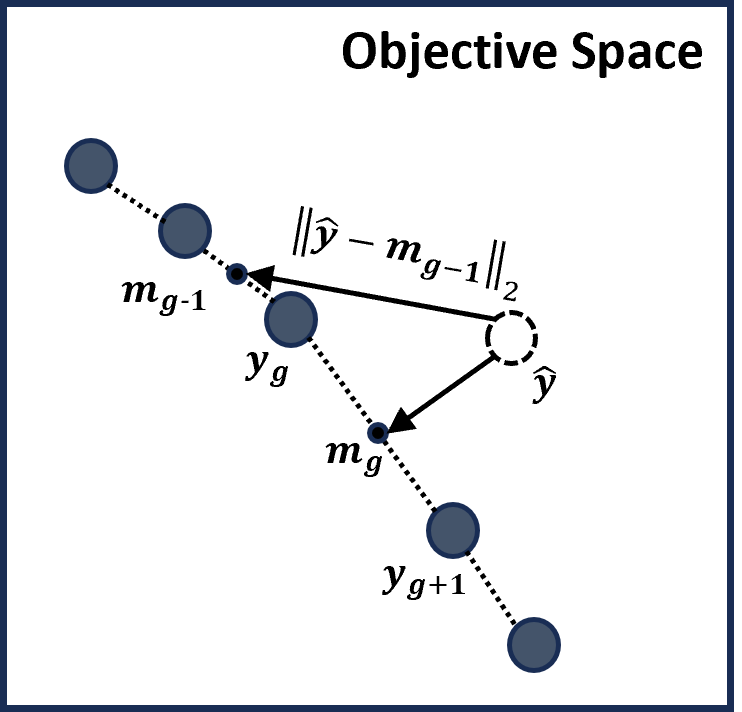}%
	}\hfill
	\subfloat[Proximity distance\label{fig:prox_distance}]{%
		\includegraphics[width=0.44\columnwidth,trim=15 20 6 20, clip]{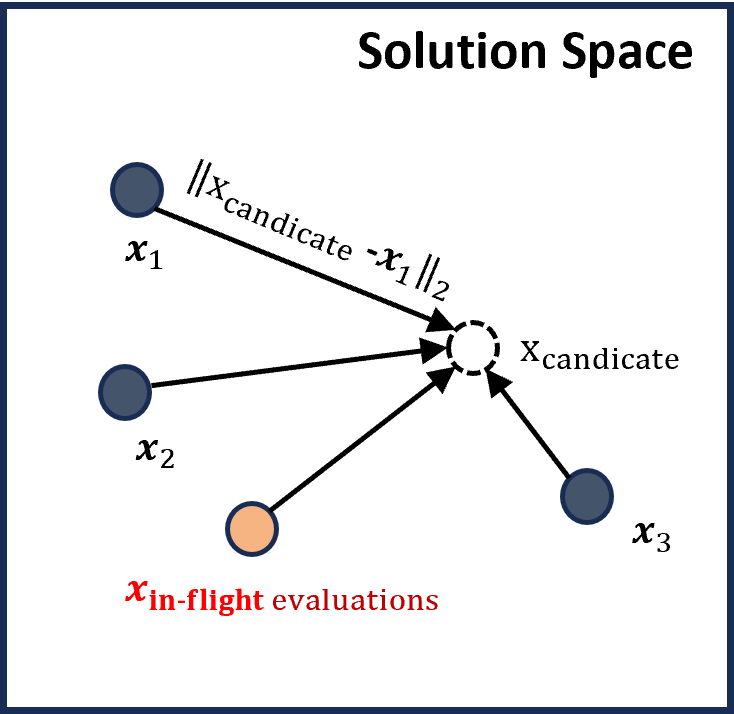}%
	}
	\caption{Illustration of distance computation for gap reward and proximity penalty.}
	\label{fig:gap_prox_distances}
\end{figure}

The gap reward is the largest weighted proximity score over all gaps:
\begin{equation}
    r_{\mathrm{gap}}(\mathbf{x}_i)
    =
    \max_{g}
    \left[
    w_g
    \exp\!\left(
    -\frac{\|\hat{\mathbf{y}}_i-\mathbf{m}_g\|_2^2}{2\sigma^2}
    \right)
    \right].
\end{equation}
Here, the exponential term is large when the predicted objective vector is close to the gap midpoint, and $w_g$ gives higher priority to wider gaps. The Gaussian kernel bandwidth $\sigma$ is computed from the overall spread of the current nondominated capability--cost front, allowing the gap reward to adapt automatically to the scale of the objective space. 
\begin{equation}
\begin{aligned}
\sigma =
\frac{1}{\max(N_{\mathrm{PF}}-1,1)}
\Bigg[
&\left(\max_g f_g^{\mathrm{cap}}-\min_g f_g^{\mathrm{cap}}\right)^2 \\
&+
\left(\max_g f_g^{\mathrm{cost}}-\min_g f_g^{\mathrm{cost}}\right)^2
\Bigg]^{1/2}.
\end{aligned}
\end{equation}
When $N_{\mathrm{PF}}<2$, no consecutive gap is available, so we set $r_{\mathrm{gap}}(\mathbf{x}_i)=0$. Thus, a candidate receives a high gap reward only when it is predicted to fill a large sparse interval in the current nondominated capability--cost front. Fig.~\ref{fig:gap_distance} illustrates the computation of gap rewards.

\paragraph{Proximity Penalty}
Candidate merge-weight vectors should avoid both currently pending evaluations and those already selected for the current dispatch. To this end, we define an avoid set $\mathcal{A}=\mathcal{P} \cup \mathcal{S}$, where $\mathcal{P}$ is the pending set, $\mathcal{S}$ contains the candidates already selected in the current dispatch. Given a candidate merge-weight vector $\mathbf{x}_i$, the proximity penalty is defined as
\begin{equation}
    p_{\mathrm{prox}}(\mathbf{x}_i;\mathcal{A})
    =
    \exp\!\left(
    k \min_{\mathbf{z}\in\mathcal{A}}\|\mathbf{x}_i-\mathbf{z}\|_2
    \right),
\end{equation}
where $k<0$ controls the decay rate of the penalty. When $\mathcal{A}=\emptyset$, we set $p_{\mathrm{prox}}(\mathbf{x}_i;\mathcal{A})=0$, since the avoid set is empty. The penalty attains its largest value for candidates closest to the avoid set and decreases exponentially as the minimum distance increases, thereby discouraging candidates that are too close to the avoid set. Fig.~\ref{fig:prox_distance} illustrates the computation of the proximity penalty.

\paragraph{Composite Ranking Score}
We combine the acquisition value, gap reward, and proximity penalty into a single ranking score:
\begin{equation}
    J(\mathbf{x}_i)
    =
    \bar{a}(\mathbf{x}_i)
    +
    \lambda_{\mathrm{gap}}\,\bar{r}_{\mathrm{gap}}(\mathbf{x}_i)
    -
    \lambda_{\mathrm{prox}}\,\bar{p}_{\mathrm{prox}}(\mathbf{x}_i;\mathcal{A}).
\end{equation}
Here, $\bar{a}$, $\bar{r}_{\mathrm{gap}}$, and $\bar{p}_{\mathrm{prox}}$ denote the acquisition value, gap reward, and proximity penalty after min-max rescaling over the current candidate pool, respectively. This rescaling places the three components on a common scale, preventing any term from dominating $J(\mathbf{x}_i)$ solely due to its numerical range. The coefficients $\lambda_{\mathrm{gap}},\lambda_{\mathrm{prox}} \ge 0$ control the strength of gap filling and duplicate avoidance.

AP-BMM selects candidates greedily according to $J$. At each selection step, it chooses the candidate with the largest ranking score among the remaining candidates:
\begin{equation}
    \mathbf{x}^\star
    =
    \arg\max_{\mathbf{x}_i\in\mathcal{C}\setminus\mathcal{S}} J(\mathbf{x}_i),
\end{equation}
where $\mathcal{C}$ is the candidate pool and $\mathcal{S}$ contains the candidates already selected in the current dispatch. The selected candidate $\mathbf{x}^\star$ is then appended to $\mathcal{S}$.

\section{Experimental Setup}
\label{sec:experimental_setup}
This section describes the experimental setup before presenting results. We separate the setup into four parts: research questions, implementation details of AP-BMM, capability--cost objective setting, baselines, evaluation metrics, and hardware platforms.

\subsection{Research Questions}
Our experiments are organized around the following three research questions.
\begin{itemize}
	\item RQ1: How does AP-BMM perform in multi-objective layer-wise model merging?
	\item RQ2: Is model-level merging sufficient to obtain a well-covered capability--cost trade-off?
	\item RQ3: Which components contribute to the performance of AP-BMM?
\end{itemize}

RQ1 evaluates AP-BMM from three complementary perspectives: Pareto-set quality, objective-space exploration, and search efficiency. RQ2 investigates whether model-level merging alone provides sufficient flexibility to achieve a well-covered capability--cost trade-off. RQ3 quantifies the contribution of the proposed prior-guided surrogate and asynchronous optimization components through ablation studies.

\subsection{Implementation Details of AP-BMM}
The Bayesian optimization component of AP-BMM consists of the discrepancy-based prior, surrogate-based acquisition optimization, and candidate ranking. For the discrepancy-based prior, we combine the parameter and activation discrepancies with $k_1=k_2=1/2$, so that neither discrepancy signal dominates before sufficient observations are collected. For surrogate-based acquisition optimization, we use BoTorch~\cite{balandat2020botorch} with $q$LogNEHVI~\cite{daulton2022lognehvi} as the acquisition function. The acquisition function is optimized with multi-start acquisition optimization~\cite{daulton2020differentiable,daulton2022multi}, using 10 restarts and 512 raw samples. Following the standard BoTorch implementation, the $q$LogNEHVI estimator uses $N=128$ Sobol quasi-Monte Carlo Gaussian samples~\cite{hou2019quasi,james1980monte} to estimate the $q$LogNEHVI acquisition value.  For candidate ranking, we set the candidate-pool multiplier to $\kappa=3$, so AP-BMM first constructs an oversized candidate pool and then ranks it to obtain the final candidates. For the proximity penalty, we set $k=-4$ to control the exponential decay rate with respect to the distance from the avoid set. We set $\lambda_{\mathrm{gap}}=0.25$ and $\lambda_{\mathrm{prox}}=0.15$ to balance gap filling and duplicate avoidance. 

Asynchronous evaluation is implemented by coupling EvalScope~\cite{evalscope_2024} with a persistent GPU worker pool. Each worker maintains a resident vLLM~\cite{kwon2025vllm} server and hot-loads merged weights through \texttt{reload\_weights}, avoiding repeated engine relaunches. This design reduces system-level overhead, so runtime differences in our experiments mainly reflect the scheduling methods.

\subsection{Experimental Settings}

\subsubsection{Capability--Cost Objective Setting}
In our experiments, we instantiate capability--cost model merging with two Qwen3-4B variants: Qwen3-4B-Thinking~\cite{yang2025qwen3} and Qwen3-4B-Instruct~\cite{yang2025qwen3}. We denote the Thinking model as $\mathcal{M}_{\mathrm{cap}}$ and the Instruct model as $\mathcal{M}_{\mathrm{cost}}$.
	
Table~\ref{table1} summarizes the objective-specific model roles and evaluation metrics. For the capability objective $f_{\mathrm{cap}}$, the Thinking model $\mathcal{M}_{\mathrm{cap}}$ serves as the expert reference because it is specialized for reasoning tasks, while the Instruct model $\mathcal{M}_{\mathrm{cost}}$ serves as the vanilla reference. For the cost objective $f_{\mathrm{cost}}$, the roles are reversed: the Instruct model $\mathcal{M}_{\mathrm{cost}}$ serves as the expert reference because it is the lower-cost model, while the Thinking model $\mathcal{M}_{\mathrm{cap}}$ serves as the vanilla reference. 

We evaluate all merged models on GPQA-Diamond~\cite{rein2024gpqa} and AIME25~\cite{aime25}, using task accuracy as the capability metric and generated-token count as the cost metric. To reduce decoding randomness, we use deterministic decoding with temperature set to $0.0$.

\begin{table}[!t]
    \caption{Objective settings for capability--cost model merging}
    \label{table1}
    \centering
    \resizebox{\columnwidth}{!}{%
    \begin{tabular}{ll}
        \toprule
    	Item & Setting \\
        \midrule
        \multicolumn{2}{l}{\textbf{Merging setup}} \\
        Source models & $\mathcal{M}_{\mathrm{cap}}$ = Thinking Model, $\mathcal{M}_{\mathrm{cost}}$ = Instruct Model \\
        \midrule
        \multicolumn{2}{l}{\textbf{Objective 1: $f_{\mathrm{cap}}$}} \\
        Expert model & $\mathcal{M}^{(\mathrm{capability})}_{\mathrm{expert}}$ = $\mathcal{M}_{\mathrm{cap}}$\\
        Vanilla model & $\mathcal{M}^{(\mathrm{capability})}_{\mathrm{vanilla}}$ = $\mathcal{M}_{\mathrm{cost}}$ \\
        Capability metrics & Accuracies on GPQA-Diamond and AIME25 \\
        \midrule
        \multicolumn{2}{l}{\textbf{Objective 2: $f_{\mathrm{cost}}$}} \\
        Expert model & $\mathcal{M}^{(\mathrm{cost})}_{\mathrm{expert}}$ = $\mathcal{M}_{\mathrm{cost}}$ \\
        Vanilla model & $\mathcal{M}^{(\mathrm{cost})}_{\mathrm{vanilla}}$ = $\mathcal{M}_{\mathrm{cap}}$ \\
		Cost metrics & Generated-token counts on GPQA-Diamond and AIME25 \\
        \bottomrule
    \end{tabular}
    }
\end{table}

\subsubsection{Baselines} 
To answer the three research questions, we design three experiments with different baselines.
For RQ1, we compare AP-BMM with four types of baselines. First, MOMM~\cite{li2025s} is included as the closest existing method, since it also addresses multi-objective layer-wise model merging. Second, we include a standard synchronous batch Bayesian optimization baseline using the $q$NEHVI acquisition function~\cite{daulton2020differentiable,daulton2022lognehvi}, denoted as $q$NEHVI. Finally, to compare with evolutionary search-based methods, we extend the single-objective TIES-DARE method~\cite{akiba2025evolutionary}, which uses CMA-ES~\cite{hansen2003reducing} to optimize layer-wise merge weights, to a multi-objective setting using MOEA/D~\cite{zhang2007moea}; we denote the resulting method as MOEA/D+CMA-ES.

All methods, i.e., AP-BMM, MOMM, $q$NEHVI, and MOEA/D+CMA-ES, use the same random seed, the same initial design of eight merge-weight vectors, and the same total budget of 40 evaluations. Thus, each method starts from the same eight initial evaluations and then performs 32 optimizer-generated evaluations, allowing us to compare optimizer behavior without confounding effects from different initial designs. Runtime is measured only for the post-initialization search phase, i.e., the 32 optimizer-generated evaluations. Notably, because LLM evaluations are expensive, the reported optimizer comparisons use a single seed; thus, the experimental results should be interpreted as budget-matched controlled comparisons under a fixed seed rather than averages over multiple seeds.

For RQ2, we compare AP-BMM with representative model-level merging methods, including TA~\cite{ilharco2022editing}, TIES-Merging~\cite{yadav2023ties}, DARE~\cite{yu2024language}, Breadcrumbs~\cite{davari2024model}, and DELLA-Merging~\cite{deep2024della}. These methods use different fusion rules, such as interpolation, sparsification, masking, and sign-based fusion, but they make a single whole-model merge decision instead of assigning layer-specific weights.

To provide a strong comparison against model-level baselines, each model-level baseline is allocated 80 evaluations, whereas AP-BMM uses 40 evaluations. TA searches over a single global merge weight, while TIES, DARE, Breadcrumbs, and DELLA search over a two-dimensional grid consisting of a global merge weight and a density parameter. The density grid is fixed to $\{0.7,0.8,0.9,1.0\}$ and crossed with 20 merge-weight values, yielding $4\times20=80$ candidates for each method. These grid-search baselines are excluded from GPU load-rate reporting because they are offline sweeps rather than online multi-GPU optimization loops; their runtime profiles are therefore not directly comparable to AP-BMM.

For RQ3, we perform an ablation study to examine the contributions of different components of AP-BMM. We compare the full AP-BMM with variants that remove the discrepancy-based prior, remove both asynchronous dispatch and pending-aware scoring, or remove only pending-aware scoring. Each ablated variant uses the same random seed, initial design, and total evaluation budget as the experiment for RQ1.

\subsubsection{Evaluation Metrics} We use Hypervolume (HV)~\cite{zitzler1998multiobjective} as the primary metric for capability--cost trade-off quality. A higher HV indicates that the discovered solution set covers a larger region of the capability--cost objective space. We also report Spacing~\cite{schott1995}, Pareto contribution ratio, and Pareto count. Spacing measures how evenly the nondominated solutions are distributed. Pareto contribution ratio measures the fraction of the pooled nondominated set contributed by each method, where the pooled nondominated set is constructed by extracting the nondominated solutions from the union of all solutions produced by all compared methods. Pareto count measures the number of nondominated solutions found within one run. 

Beyond scalar metrics, we visualize task accuracy versus generated-token count  on AIME25 and GPQA-Diamond to examine whether an optimizer covers a broad range of trade-offs or collapses toward one extreme.

\subsubsection{Hardware Platform:} All experiments are conducted on a single workstation with four NVIDIA GeForce RTX 4090 GPUs, each with 24 GB of memory, and two Intel Xeon Platinum 8336C CPUs, each with 32 physical cores. The workstation has approximately 512 GiB of system memory.

\section{Results and Analysis}
\label{sec:main_results}

\subsection{RQ1: Comparison with Layer-wise Merging Methods}


\subsubsection{Overall Pareto Performance}
Table~\ref{tab:main_results} compares AP-BMM with other layer-wise model merging methods. AP-BMM achieves the best overall performance, with the highest HV of 1.3338 and the lowest spacing of 0.0216. This indicates that AP-BMM not only covers a larger capability--cost objective-space volume, but also produces a more evenly distributed Pareto set. AP-BMM also contributes 7 out of 11 solutions to the pooled nondominated set, showing that most competitive Pareto solutions are produced by our method.
\begin{table}[!t]
	\caption{Comparison of AP-BMM with other layer-wise model merging methods.}
	\label{tab:main_results}
	\centering
	\resizebox{\columnwidth}{!}{%
		\begin{tabular}{lccccc}
			\toprule
			Method & Category & HV$\uparrow$ & Spacing$\downarrow$ & Contrib.$\uparrow$ & \#Pareto Sol.$\uparrow$ \\
			\midrule
			\textbf{AP-BMM (ours)} & Bayesian & \textbf{1.3338} & \textbf{0.0216} & \textbf{7/11} & \textbf{9} \\
			MOMM  & Bayesian & 1.1665     & 0.0859     & 3/11     & \textbf{9} \\
			$q$NEHVI & Bayesian & 1.1298 & 0.0551 & 0/11 & 6 \\
			MOEA/D+CMA-ES & Evolutionary & 1.1707 & 0.1209 & 1/11 & 6 \\
			\bottomrule
		\end{tabular}
	}
\end{table}

Among the baselines, MOEA/D+CMA-ES obtains the second-best HV, while MOMM finds the same number of internal Pareto solutions as AP-BMM. However, both methods have worse spacing and lower contribution to the pooled nondominated set, suggesting that their solutions are either less evenly distributed or more often dominated by AP-BMM. The $q$NEHVI baseline has relatively good spacing but contributes no solution to the pooled nondominated set, indicating that its discovered trade-offs are dominated when compared with the other methods. Overall, the results show that AP-BMM provides a stronger and better-covered approximation to the capability--cost Pareto set than existing layer-wise merging methods.

\subsubsection{Objective-Space Distribution Analysis}

\begin{figure}[!t]
	\begin{center}
		\includegraphics[width=0.99\columnwidth]{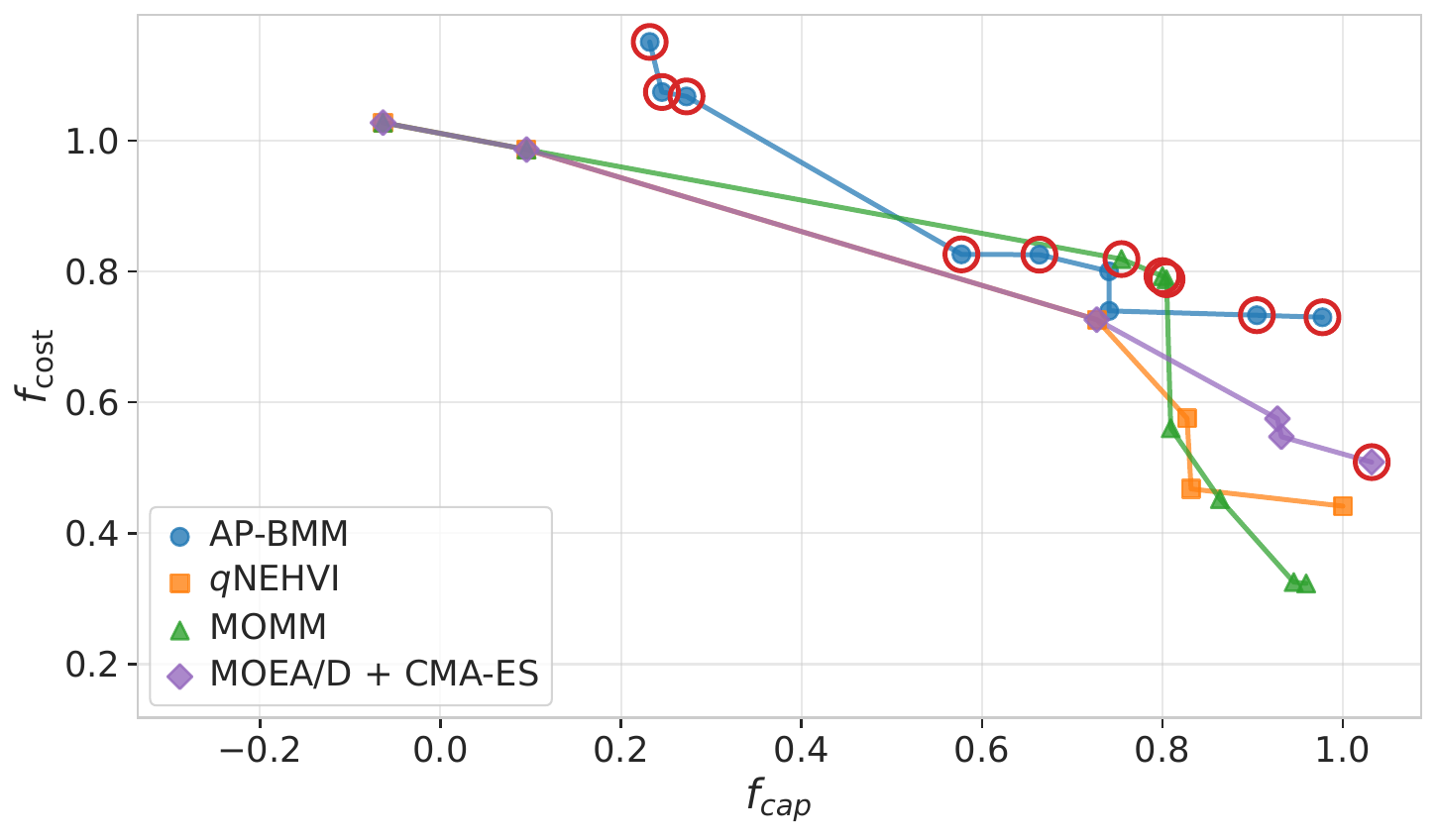}
		\caption{Pareto fronts found by AP-BMM and other layer-wise merging methods. Red circles mark pooled nondominated solutions.}
		\label{fig:pareto_front_2d}
	\end{center}
\end{figure}

Fig.~\ref{fig:pareto_front_2d} visualizes the Pareto fronts found by AP-BMM and other layer-wise merging methods in the capability--efficiency objective space. AP-BMM produces a broader and more evenly distributed Pareto front than the baselines. Its solutions cover high-efficiency regions at moderate capability as well as high-capability regions with reasonable efficiency, forming a smoother trade-off across the objective space. This indicates that AP-BMM can discover diverse layer-wise merge weights instead of concentrating on a narrow portion of the front.

Compared with the baselines, $q$NEHVI mainly finds solutions in the high-capability but lower-efficiency region and misses much of the middle- and high-efficiency front. MOMM includes some competitive solutions near the middle of the trade-off, but its front drops sharply in efficiency as capability increases, leading to weaker coverage at the high-capability end. MOEA/D+CMA-ES covers a broader range than $q$NEHVI, but it is still dominated by AP-BMM over much of the front. The red-circled points indicate solutions that remain in the pooled nondominated set; most of these are produced by AP-BMM. This visual evidence is consistent with Table~\ref{tab:main_results}, where AP-BMM achieves the highest HV, the best spacing, and the largest contribution to the pooled nondominated set.

\begin{figure}[!t]
	\centering
	\subfloat[AIME25]{
	\begin{minipage}[t]{\columnwidth}
		\centering
		\includegraphics[width=\linewidth]{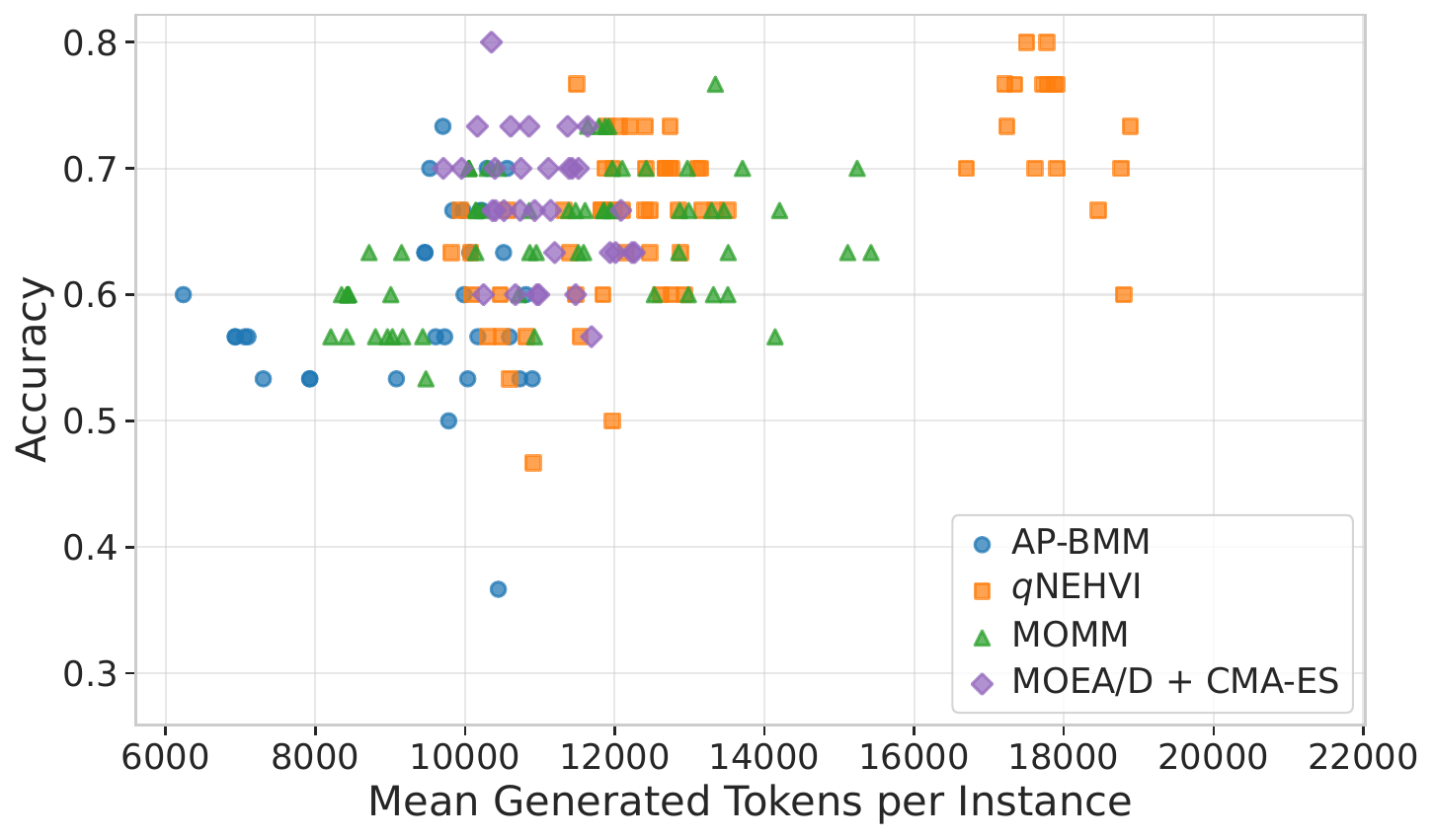}
	\end{minipage}
	}
	\hfill
	\subfloat[GPQA-Diamond]{
	\begin{minipage}[t]{\columnwidth}
		\centering
		\includegraphics[width=0.9\linewidth]{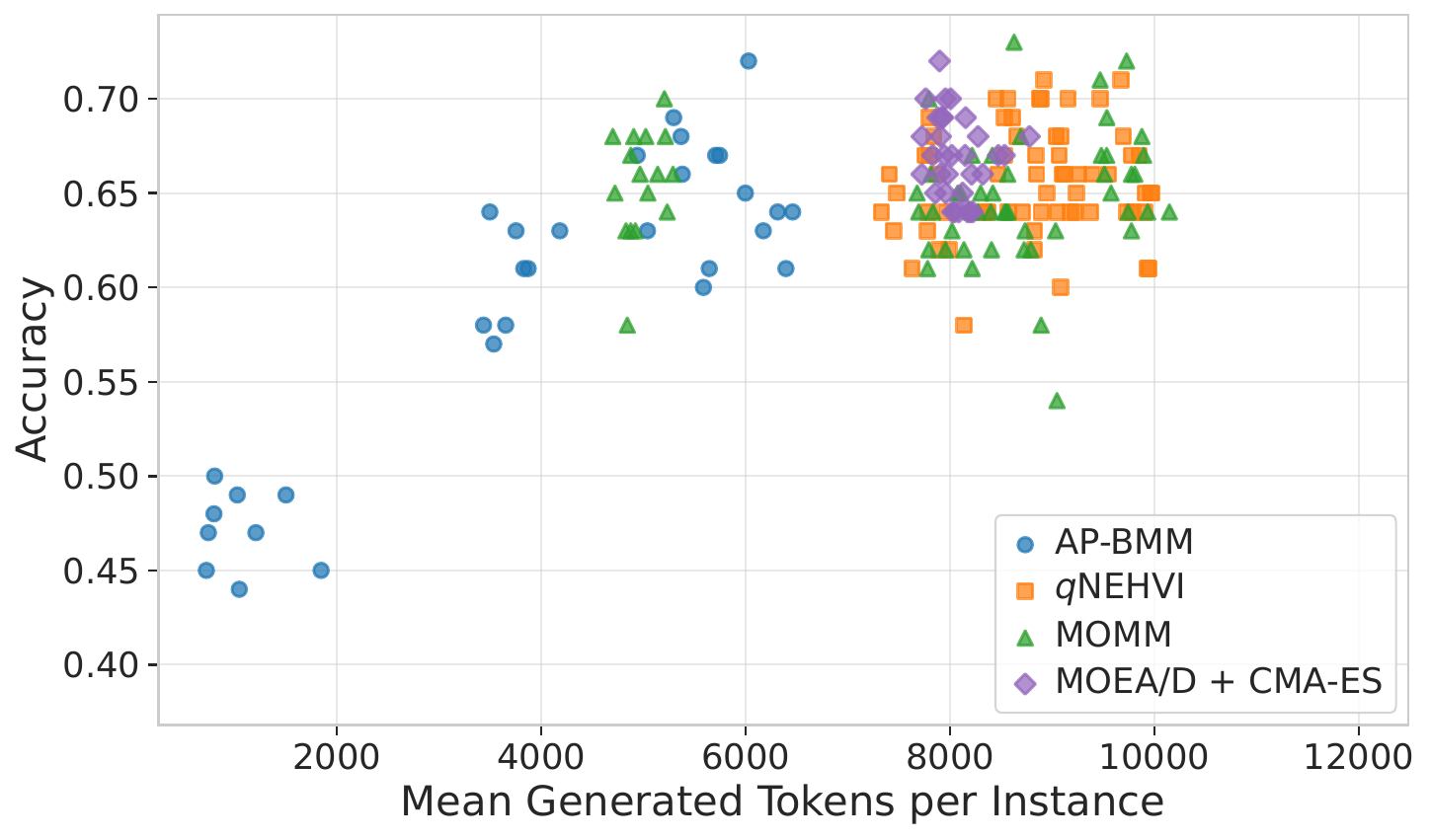}
	\end{minipage}
	}
	\caption{Accuracy vs. generated-token count of AP-BMM and other layer-wise merging methods on AIME25 and GPQA-Diamond.}
	\label{fig:solution_views_main}
\end{figure}

To show where each method spends its evaluation budget, Fig.~\ref{fig:solution_views_main} provides a full view of all evaluated merged models in the accuracy--token-count space on AIME25 and GPQA-Diamond. Each point represents one merged model, with accuracy on the y-axis and the mean number of generated tokens per instance on the x-axis. Across both datasets, AP-BMM produces a broader spread across generated-token counts, while the baselines tend to concentrate in narrower regions, especially at higher token counts on GPQA-Diamond. This indicates that AP-BMM explores the capability--cost space more effectively and can identify lower-cost merged models without sacrificing competitive task performance.

\subsubsection{Search Efficiency}

We analyze search efficiency from both algorithmic and system-level perspectives. The former measures how effectively each method converts completed evaluations into Pareto-set improvement, while the latter measures how efficiently these evaluations are scheduled on the shared GPU worker pool.

\begin{figure}[!t]
  \centering
  \subfloat[HV Value.\label{fig:hvvalues}]{
  \begin{minipage}{\columnwidth}
   \centering
   \hspace{-4mm}
   \includegraphics[width=0.85\linewidth]{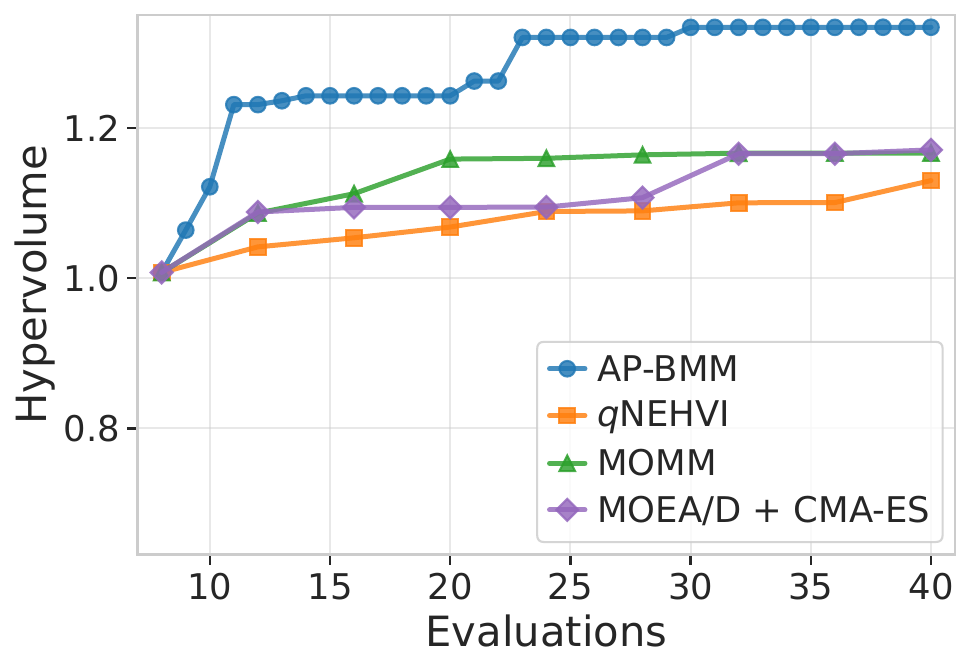}
  \end{minipage}%
  }
  \hfill
  \subfloat[Pareto set size\label{fig:pareto_count}]{
  \begin{minipage}{\columnwidth}
   \centering
   \includegraphics[width=0.82\linewidth]{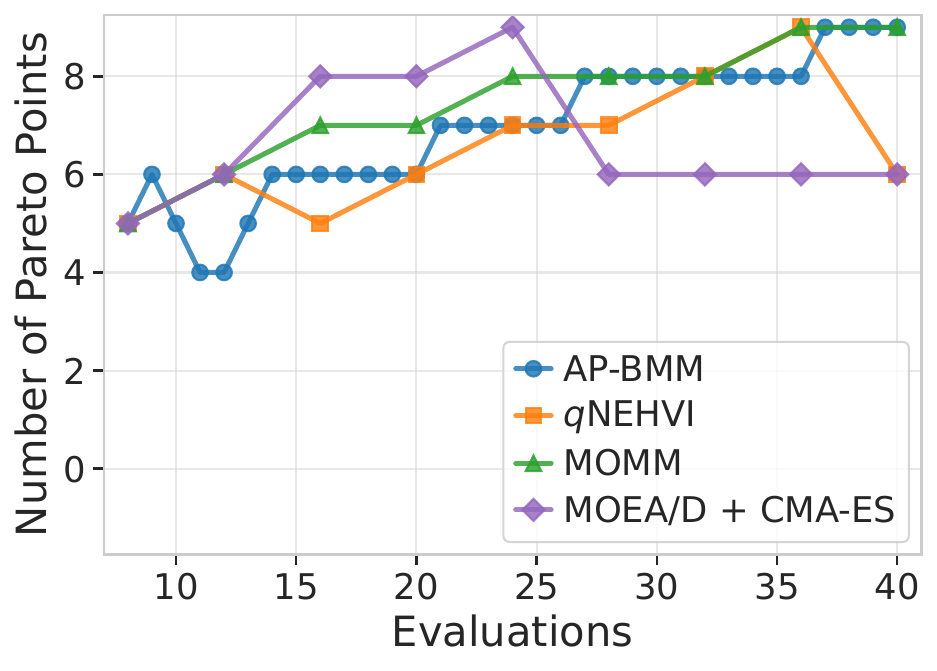}
  \end{minipage}
  }
  \caption{Search progress of AP-BMM and other layer-wise merging methods. The upper panel reports the HV trajectory, and the lower panel reports the growth of the discovered Pareto set. The x-axis denotes the number of completed evaluations.}
  \label{fig:hv_iteration_main}
 \end{figure}

Fig.~\ref{fig:hv_iteration_main} shows the algorithmic search progress over completed evaluations after the shared initialization phase, with the upper panel illustrating the HV trajectory and the lower panel reporting the growth of the discovered Pareto set size. AP-BMM increases HV rapidly and reaches a high plateau earlier than the baselines, indicating that it identifies useful capability--cost trade-offs with fewer completed evaluations. In contrast, MOMM and $q$NEHVI exhibit slower HV improvement, while MOEA/D+CMA-ES exhibits larger fluctuations in the number of discovered Pareto points. These trends suggest that AP-BMM's advantage is not limited to the final Pareto set, but also appears throughout the search progress. This behavior is consistent with the intended effect of accounting for pending candidates and ranking new candidates to reduce duplicate exploration.

\begin{figure}[!t]
	\begin{center}
		\centerline{\includegraphics[width=0.98\columnwidth]{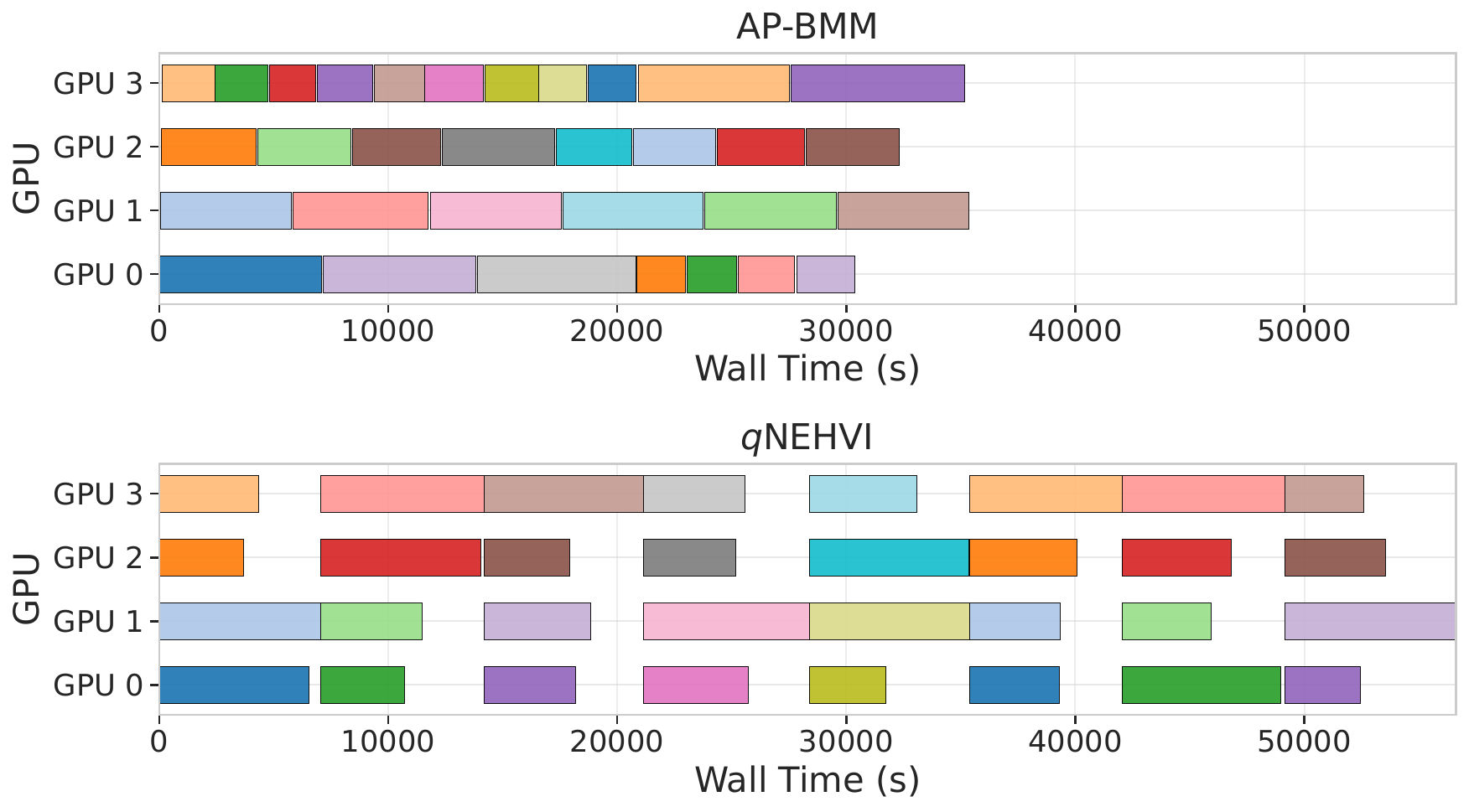}}
		\caption{GPU utilization timelines of AP-BMM and $q$NEHVI.}
		\label{fig:gantt_apbmm_qnehvi}
	\end{center}
\end{figure}

Beyond algorithmic efficiency, AP-BMM also improves wall-clock efficiency through asynchronous scheduling. Fig.~\ref{fig:gantt_apbmm_qnehvi} visualizes the resulting scheduling difference between AP-BMM and the synchronous Bayesian baseline $q$NEHVI~\cite{daulton2020differentiable,daulton2022lognehvi}. AP-BMM keeps the 4-worker pool active by dispatching a new candidate as soon as any GPU becomes idle, leading to the denser timeline in the AP-BMM panel. In contrast, $q$NEHVI must wait for the slowest task in each batch before updating the surrogate and launching the next round, which creates wider idle gaps. Over the shared 32-task post-initialization window, AP-BMM increased the GPU load rate from 73.17\% to 92.57\% and reduced runtime from 15.73h to 9.84h. Notably, since both methods use the same worker pool and inference stack, this comparison isolates the system-level effect of asynchronous versus synchronous Bayesian search.

\subsection{RQ2: Comparison with Model-level Merging Methods}

\subsubsection{Quantitative Pareto-set Comparison}
Table~\ref{tab:granularity_summary} compares AP-BMM with representative model-level merging methods. AP-BMM achieves the highest HV, 1.3338, and the lowest nonzero spacing, 0.0216, indicating that layer-wise Bayesian merging finds both stronger and more evenly distributed capability--cost trade-offs. It also contributes 9 out of 15 solutions to the pooled nondominated set, substantially more than any model-level baseline, showing that most of the final Pareto front is provided by AP-BMM.

\begin{table}[!t]
	\caption{Comparison of AP-BMM with model-level merging methods.}
	\label{tab:granularity_summary}
	\begin{center}
		\resizebox{\columnwidth}{!}{%
			\begin{footnotesize}
				\setlength{\tabcolsep}{3pt}
					\begin{tabular}{lcccc}
						\toprule
						Method & HV $\uparrow$ & Spacing $\downarrow$ & Contrib. $\uparrow$ & \#Pareto Sol.  $\uparrow$ \\
						\midrule
						\textbf{AP-BMM (ours)} & \textbf{1.3338} & \textbf{0.0216} & \textbf{9/15} & \textbf{9} \\
						Breadcrumbs & 1.0862 & 0.1342 & 2/15 & 7 \\
						TA & 0.7092 & 0.1013 & 1/15 & 4 \\
						DARE & 0.6651 & 0.0000 & 1/15 & 2 \\
						DELLA-Merging & 0.3847 & 0.0000 & 0/15 & 2 \\
						TIES-Merging & 0.3614 & 0.0467 & 2/15 & 3 \\
						\bottomrule
					\end{tabular}
			\end{footnotesize}
		}
	\end{center}
\end{table}

Among the model-level methods, Breadcrumbs obtains the strongest HV, 1.0862, but still lags behind AP-BMM and has worse spacing, suggesting less uniform coverage. TA, DARE, DELLA-Merging, and TIES-Merging produce lower HV and fewer competitive Pareto solutions. The zero spacing values for DARE and DELLA-Merging arise because they return very few nondominated solutions, making the spacing metric less informative. Overall, these results show that model-level merging is insufficient for well-covered capability--cost optimization, while AP-BMM benefits from the finer control provided by layer-wise merge weights.

\subsubsection{Objective-space Distribution Analysis}
\begin{figure}[!t]
	\centering
	\begin{minipage}[t]{\columnwidth}
		\centering
		\includegraphics[width=0.9\linewidth]{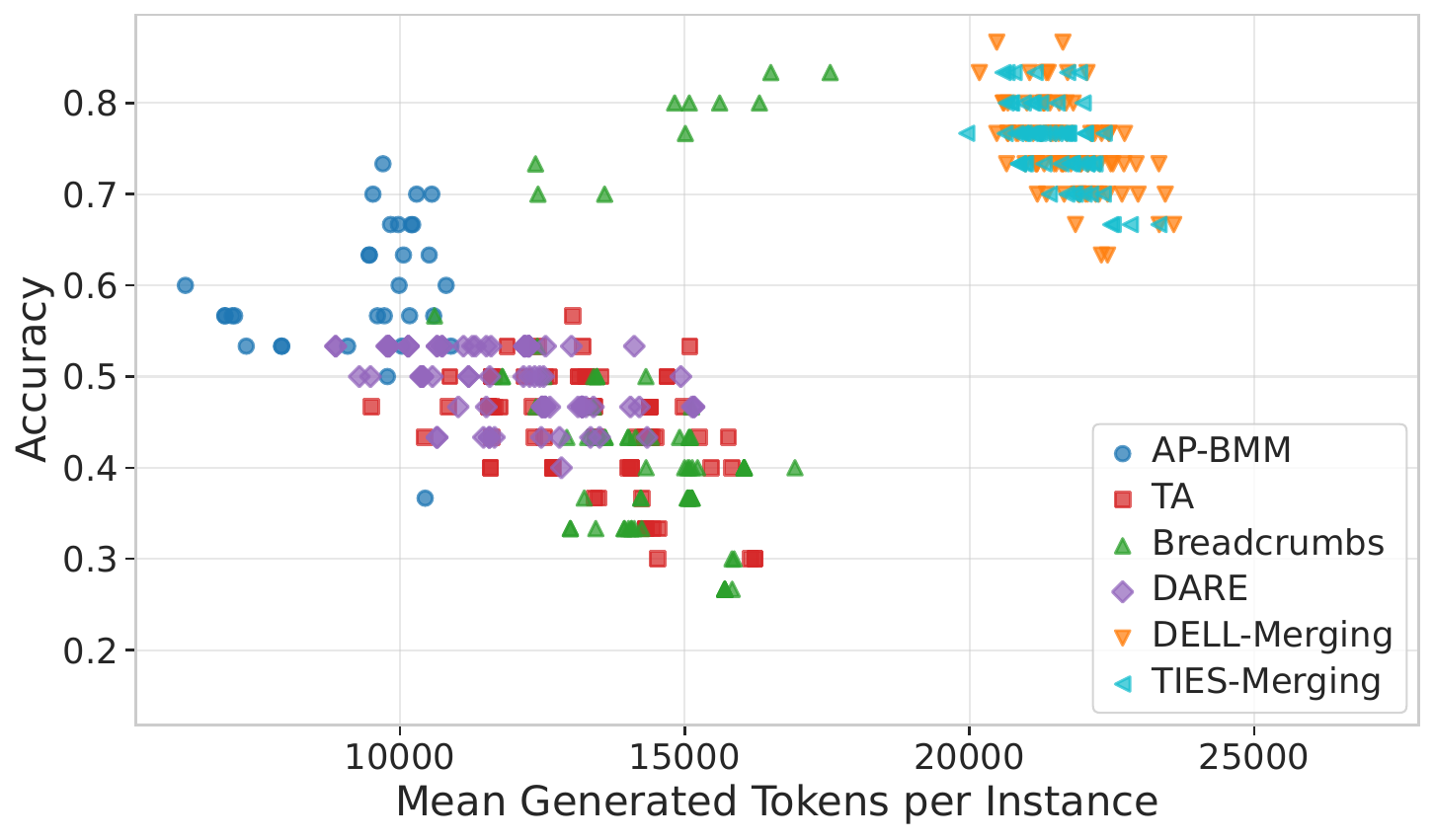}
		
		(a) AIME25
	\end{minipage}
	\begin{minipage}[t]{\columnwidth}
		\centering
		\includegraphics[width=0.9\linewidth]{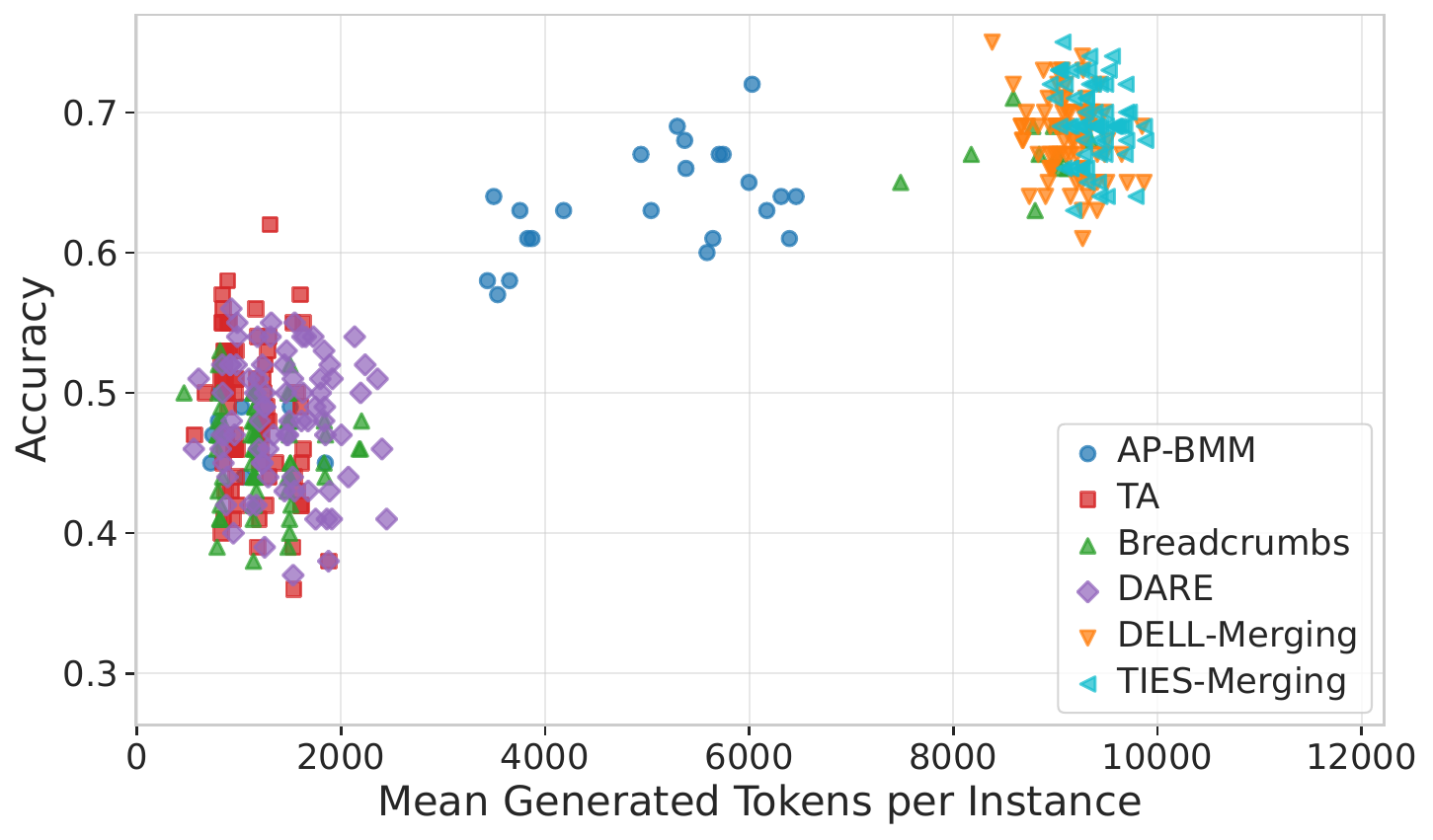}
		
		(b) GPQA-Diamond
	\end{minipage}
	\caption{Accuracy vs. generated-token count of AP-BMM and model-level merging methods on AIME25 and GPQA-Diamond.}
	\label{fig:solution_views_granularity}
\end{figure}

Fig.~\ref{fig:solution_views_granularity} visualizes the capability--cost trade-offs found by AP-BMM and model-level merging methods on AIME25 and GPQA-Diamond. On both datasets, AP-BMM produces solutions that span a wider range of token counts while maintaining competitive accuracy, indicating that layer-wise search can explore more diverse capability--cost merge weights than model-level merging, even under a doubled evaluation budget for the baselines.

On AIME25, AP-BMM discovers solutions across the low- and mid-token-count regions and includes several high-accuracy solutions. In contrast, TA and DARE tend to favor shorter outputs but remain in lower-accuracy regions, while high-accuracy solutions from model-level baselines often require substantially more generated tokens. DELLA-Merging and TIES-Merging, in particular, cluster near the high-token-count region, suggesting limited ability to reduce generation cost while preserving capability. On GPQA-Diamond, AP-BMM again provides a broader spread across token counts, including competitive accuracy points at substantially lower token counts than the high-token-count clusters of DELLA-Merging and TIES-Merging. These patterns show that model-level merging methods tend to produce narrower clusters of solutions, whereas AP-BMM's layer-wise optimization better explores the capability--cost space and yields more useful candidates.

\subsection{RQ3: Ablation Study of AP-BMM Components}
Table~\ref{tab:ablation_metrics} reports the ablation study of AP-BMM. Full AP-BMM uses all proposed components. w/o Prior removes the layer-wise prior. w/o Async+Pending replaces the asynchronous loop with synchronous batch optimization and also removes the pending-aware acquisition. w/o Pending keeps asynchronous dispatch but scores new candidates without accounting for candidates already under evaluation.

\begin{table}[!t]
	\caption{Ablation study of AP-BMM components.}
	\label{tab:ablation_metrics}
	\begin{center}
		\resizebox{\columnwidth}{!}{%
			\begin{footnotesize}
				\setlength{\tabcolsep}{4pt}
					\begin{tabular}{lcccc}
						\toprule
						Method & HV $\uparrow$ & Spacing $\downarrow$ & Contrib. $\uparrow$ & \#Pareto Sol. $\uparrow$ \\
						\midrule
						\textbf{Full AP-BMM} & \textbf{1.3338} & \textbf{0.0216} & \textbf{9/12} & 9 \\
						w/o Prior & 1.1479 & 0.0881 & 0/12 & \textbf{12} \\
						w/o Async+Pending & 1.1832 & 0.2348 & 1/12 & 7 \\
						w/o Pending & 1.2630 & 0.1278 & 2/12 & 11 \\
						\bottomrule
					\end{tabular}
			\end{footnotesize}
		}
	\end{center}
\end{table}

Full AP-BMM achieves the best HV and spacing, showing that it finds a stronger capability--cost Pareto set with the most uniform coverage. Removing the prior decreases HV from 1.3338 to 1.1479 and worsens spacing from 0.0216 to 0.0881, indicating that discrepancy-guided lengthscale initialization helps the GP focus the search on informative layer dimensions. Although w/o Prior produces the largest number of internal Pareto solutions, its contribution is 0/12, meaning that none of these solutions remains nondominated when compared with solutions from other variants.

The asynchronous and pending-aware components address different sources of inefficiency. The w/o Async+Pending variant obtains the worst spacing score, 0.2348, showing that removing the asynchronous dispatch and pending-aware acquisition leads to poorly distributed solutions. Comparing w/o Pending with w/o Async+Pending isolates the benefit of asynchronous dispatch. In synchronous batch optimization, all candidates in a batch are selected from the same surrogate state; if early completed evaluations reveal that a region is less promising, the remaining candidates in the batch have already been dispatched and cannot be adjusted. Asynchronous dispatch avoids this rigid batch commitment by allowing each returned evaluation to influence subsequent candidate selection, which explains why w/o Pending improves over w/o Async+Pending. However, w/o Pending still underperforms Full AP-BMM in both HV and spacing because it does not account for candidates already under evaluation. The pending-aware acquisition corrects this by including pending candidates when estimating hypervolume contribution, reducing redundant candidates and improving coverage. Overall, the ablation confirms that prior guidance, asynchronous scheduling, and pending-aware acquisition are complementary for improving Pareto-set quality.

\section{Conclusion}
\label{sec:conclusion}

This paper presented AP-BMM, an asynchronous prior-guided Bayesian optimization framework for multi-objective layer-wise model merging. AP-BMM formulates layer-wise merge-weight optimization as a capability--cost Pareto optimization problem and combines prior-guided Gaussian process surrogates with asynchronous, pending-aware candidate selection to improve both optimization quality and search efficiency. Compared with existing layer-wise and model-level merging methods, AP-BMM discovers a broader and better-distributed capability--cost Pareto set while significantly improving GPU utilization through asynchronous evaluation. Extensive experiments demonstrate that AP-BMM consistently achieves superior Pareto-set quality, explores a wider range of capability--cost trade-offs, and benefits from the complementary effects of prior-guided surrogate modeling, asynchronous scheduling, and pending-aware acquisition.

There are several promising directions for future work. First, the current framework focuses on merging two source models; extending AP-BMM to efficiently optimize layer-wise merging across multiple source models would further increase its applicability. Second, richer prior knowledge, such as parameter similarity or representation-level information, could be incorporated to improve surrogate modeling and search efficiency. Finally, applying the proposed asynchronous multi-objective optimization framework to broader model customization tasks, such as continual adaptation and multi-task model composition, is another interesting direction.

\bibliographystyle{IEEEtran}
\bibliography{ref}

%% file: supplementary2.tex
	\title{Supplementary Material for ``AP-BMM: Approximating Capability-Cost Pareto Sets of LLMs via Asynchronous Prior-Guided Bayesian Model Merging''}
	
	\author{Kesheng~Chen, Yamin~Hu, Zhenqian~Zhu, Yiya~Diao, Wenjian~Luo,~\IEEEmembership{Senior Member,~IEEE}%
		\thanks{Kesheng~Chen, Yamin~Hu, Zhenqian~Zhu, Yiya~Diao and Wenjian~Luo are with Guangdong Provincial Key Laboratory of Novel Security Intelligence Technologies, Institute of Cyberspace Security, School of Computer Science and Technology, Harbin Institute of Technology, Shenzhen 518055, China. (e-mail: 22s151138@stu.hit.edu.cn, huyamin@hit.edu.cn, 23b351010@stu.hit.edu.cn, diaoyiyacug@gmail.com, luowenjian@hit.edu.cn).}
	}

	\maketitle

	\tableofcontents

	\noindent\textbf{Supplementary overview.}
	This document provides implementation and analysis details that support the main paper. Section~\ref{app:baseline_details} defines the model-level merging baselines used in the model-level comparison. Section~\ref{app:moead_cmaes_details} describes the synchronous MOEA/D + CMA-ES layer-wise baseline. Section~\ref{app:prior_visualization} visualizes the discrepancy-based layer-wise prior used by AP-BMM. Section~\ref{app:pooled_pareto_solutions} reports representative pooled best trade-off solutions, and Section~\ref{app:qualitative_examples} provides qualitative responses from representative evaluated merged models. The terminology follows the main paper: the optimization objective is a capability--cost trade-off, generated-token count is used as the primary cost proxy, and all objective values used for Pareto-quality metrics are reference-scaled.

	\section{Model-level Merging Baselines}
	\label{app:baseline_details}
	
	This section provides the mathematical definitions of the model-level baselines used in the model-level comparison of the main paper. These baselines operate at the whole-model level: after constructing task vectors relative to a common base model, they apply one global fusion coefficient and, where applicable, one global density or masking hyperparameter across the entire model. This differs from AP-BMM, which searches for layer-wise merge weights.
	
	Let $\mathcal{M}_{\text{base}}$ be the base model and $\mathcal{M}^{(i)}$ be the $i$-th task model. We use $N_s$ to denote the number of task models, reserving $K$ for the number of optimization objectives as in the main paper. The task vector of the $i$-th task model is
	\[
	\TV^{(i)} := \mathcal{M}^{(i)} - \mathcal{M}_{\text{base}}.
	\]
	
	\subsection{Task Arithmetic}
	Task Arithmetic~\cite{ilharco2022editing} obtains a merged model by adding a weighted sum of task vectors to the base model:
	\begin{equation}
		\mathcal{M}_{\text{merge}}
		=
		\mathcal{M}_{\text{base}}
		+
		\alpha \sum_{i=1}^{N_s} \TV^{(i)}.
	\end{equation}
	Here $\alpha$ is a global fusion coefficient shared across all layers, consistent with the model-level formulation in the main paper.
	
	\subsection{TIES-Merging}
	TIES~\cite{yadav2023ties} addresses parameter conflicts using sparsification, sign resolution, and selective merging. A simplified whole-model form can be written as
	\begin{equation}
		\mathcal{M}_{\text{merge}}
		=
		\mathcal{M}_{\text{base}}
		+
		\alpha \sum_{i=1}^{N_s} \mathrm{TIES}\!\left(\TV^{(i)}\right),
	\end{equation}
	where $\mathrm{TIES}(\cdot)$ denotes the sparsification, sign-resolution, and selective-merging operator used by TIES.
	
	\subsection{DARE-Merging}
	DARE (Drop And REscale)~\cite{yu2024language} randomly drops delta parameters and rescales the remaining entries. Writing $\rho$ for the retained density, its model-level form can be expressed as
	\begin{equation}
		\mathcal{M}_{\text{merge}}
		=
		\mathcal{M}_{\text{base}}
		+
		\alpha \sum_{i=1}^{N_s} \mathrm{DARE}\!\left(\TV^{(i)}, \rho\right).
	\end{equation}
	In the main experiments, $\alpha$ and the retained density $\rho$ are searched on a grid for the model-level comparison.
	
	\subsection{Breadcrumbs}
	Breadcrumbs~\cite{davari2024model} applies sparsification and masking to remove both large outliers and small perturbations:
	\begin{equation}
		\mathcal{M}_{\text{merge}}
		=
		\mathcal{M}_{\text{base}}
		+
		\alpha \sum_{i=1}^{N_s} m^{(i)} \odot \TV^{(i)},
	\end{equation}
	where $m^{(i)}$ is a mask and $\odot$ denotes element-wise multiplication.
	
	\subsection{DELLA}
	DELLA (Drop--Elect--Fuse)~\cite{deep2024della} drops parameters, elects deltas with consistent signs, and fuses the selected deltas:
	\begin{equation}
		\mathcal{M}_{\text{merge}}
		=
		\mathcal{M}_{\text{base}}
		+
		\alpha \cdot \delta^{(\mathrm{avg})},
	\end{equation}
	where $\delta^{(\mathrm{avg})}$ is the averaged and rescaled delta after the Drop--Elect--Fuse pipeline, and $\alpha$ is the global fusion coefficient.
	
	\section{Implementation Details of MOEA/D + CMA-ES}
	\label{app:moead_cmaes_details}
	
	This section details the MOEA/D~\cite{zhang2007moea} + CMA-ES~\cite{hansen2003reducing} baseline used in the layer-wise optimization comparison. The baseline uses the same layer-wise decision vector, reference-scaled objectives, initial design, and evaluation budget as AP-BMM. The key difference is the optimization loop: MOEA/D + CMA-ES remains synchronous and updates only after a full batch has finished, whereas AP-BMM updates after each worker returns an evaluation result.
	
	\subsection{MOEA/D Decomposition and Neighborhood Structure}
	MOEA/D~\cite{zhang2007moea} maintains a set of scalar subproblems indexed by weight vectors $\{\boldsymbol{\lambda}_i\}_{i=1}^{M}$, with $\boldsymbol{\lambda}_i\in \mathbb{R}_+^2$ and $\|\boldsymbol{\lambda}_i\|_1=1$. Each subproblem corresponds to one scalar aggregation of the two reference-scaled objectives. For each subproblem $i$, the neighbor set $\mathcal{N}(i)$ consists of the $T_{\mathrm{nb}}$ closest weight vectors under Euclidean distance.
	
	For intuition, the decomposition can be viewed as minimizing the distance to the current ideal point $\mathbf{z}$ under a Tchebycheff-style scalarization:
	\begin{equation}
		g(\mathbf{x}\mid \boldsymbol{\lambda}_i,\mathbf{z})
		=
		\max_{k\in\{1,2\}}
		\lambda_{i,k}\,\big| f_k(\mathbf{x}) - z_k \big|.
	\end{equation}
	Since the main paper casts both objectives into a maximize-higher-is-better form, $\mathbf{z}$ is updated as the componentwise best objective value observed so far.
	
	\subsection{Population Decomposition and Neighborhoods}
	Let $q$ denote the synchronous batch size. The number of MOEA/D subproblems is set as
	\[
	M=\max(8,\min(24,4q)).
	\]
	With two objectives and $q=4$, this yields $M=16$ decomposition weights uniformly spaced on the unit simplex. For each subproblem $i$, the algorithm constructs a neighborhood $\mathcal{N}(i)$ from the nearest weight vectors. These neighbors define the local region from which incumbent solutions are borrowed and provide differential information for perturbing new CMA-style samples.
	
	Each subproblem keeps its own local optimizer state:
	\[
	\{\mathbf{m}_i,\sigma_i,\mathbf{C}_i,\mathbf{p}^{(\sigma)}_i,\mathbf{p}^{(c)}_i\},
	\]
	where $\mathbf{m}_i$ is the latent mean, $\sigma_i$ is the global step size, $\mathbf{C}_i$ is the covariance matrix, and $\mathbf{p}^{(\sigma)}_i,\mathbf{p}^{(c)}_i$ are the standard CMA-ES~\cite{hansen2003reducing} evolution paths. The implementation also stores the current incumbent solution and scalarized score for each subproblem.
	
	\subsection{Candidate Generation}
	For each active subproblem $i$, the algorithm first estimates a local center by combining the current mean with the average incumbent of valid neighboring subproblems. It then samples offspring from a Gaussian search distribution:
	\begin{equation}
		\mathbf{u}_{i,j}
		=
		\mathbf{m}_i + \sigma_i \mathbf{L}_i \boldsymbol{\epsilon}_{i,j},
		\qquad
		\boldsymbol{\epsilon}_{i,j}\sim\mathcal{N}(\mathbf{0},\mathbf{I}),
	\end{equation}
	where $\mathbf{L}_i\mathbf{L}_i^\top=\mathbf{C}_i$. If at least two valid neighboring incumbents exist, the implementation additionally injects a differential perturbation
	\[
	\mathbf{u}_{i,j}
	\leftarrow
	\mathbf{u}_{i,j} + 0.15\left(\mathbf{u}^{\star}_{a}-\mathbf{u}^{\star}_{b}\right),
	\]
	with $a,b\in \mathcal{N}(i)$ sampled from the valid-neighbor set, and $\mathbf{u}^{\star}_{a}$ and $\mathbf{u}^{\star}_{b}$ denoting incumbent latent vectors. This term is not part of textbook CMA-ES~\cite{hansen2003reducing}; it is an implementation detail that mixes MOEA/D neighborhood structure with a lightweight differential-evolution-style directional bias. The sampled latent vectors are clipped to $[-1,1]^{D'}$, decoded back to layer-wise merge coefficients, and evaluated by the same model-building and benchmark pipeline as AP-BMM. Here $D'$ is the internal search dimension, equal to the number of optimized layer-wise or block-wise coefficients used by the baseline.
	
	\subsection{Selection and CMA-ES-Style State Updates}
	After the synchronous batch finishes, each offspring is scored against the relevant decomposition weights. The implementation uses an archive-rescaled Tchebycheff criterion together with a small sum penalty:
	\begin{equation}
		\begin{aligned}
			s(\mathbf{x};\boldsymbol{\lambda}_i)
			=
			-\Big(
			&\max_k \lambda_{i,k}\!\left(1-\tilde{f}_k(\mathbf{x})\right) \\
			&+ 0.05\sum_k \lambda_{i,k}\!\left(1-\tilde{f}_k(\mathbf{x})\right)
			\Big),
		\end{aligned}
	\end{equation}
	where $\tilde{f}_k(\mathbf{x})$ is the archive-rescaled objective value computed from the observed candidates. This rescaling is used only inside the evolutionary baseline's scalarized selection rule and is separate from the reference-scaled objective definition in the main paper. Because the score is negated, a larger $s(\mathbf{x};\boldsymbol{\lambda}_i)$ is better. The best observed candidate for each subproblem is updated whenever this score improves.
	
	For the active subproblems, the local CMA state is then updated from the elite offspring. The new mean is formed by weighted recombination of the top-ranked local samples, optionally blended with the average incumbent of the neighborhood. The evolution paths $\mathbf{p}^{(\sigma)}_i$ and $\mathbf{p}^{(c)}_i$, the covariance $\mathbf{C}_i$, and the step size $\sigma_i$ are updated with standard CMA-ES-style equations. The implementation slightly enlarges $\sigma_i$ when no local scalarized improvement is observed, which helps prevent early stagnation. After the state update, the algorithm recomputes the $\HV$, appends the new batch to the evaluation archive, and begins the next synchronous iteration.

	\section{Layer-wise Prior Visualization}
	\label{app:prior_visualization}
	
\begin{figure}[htbp]
		\centering
		\includegraphics[width=0.4\textwidth]{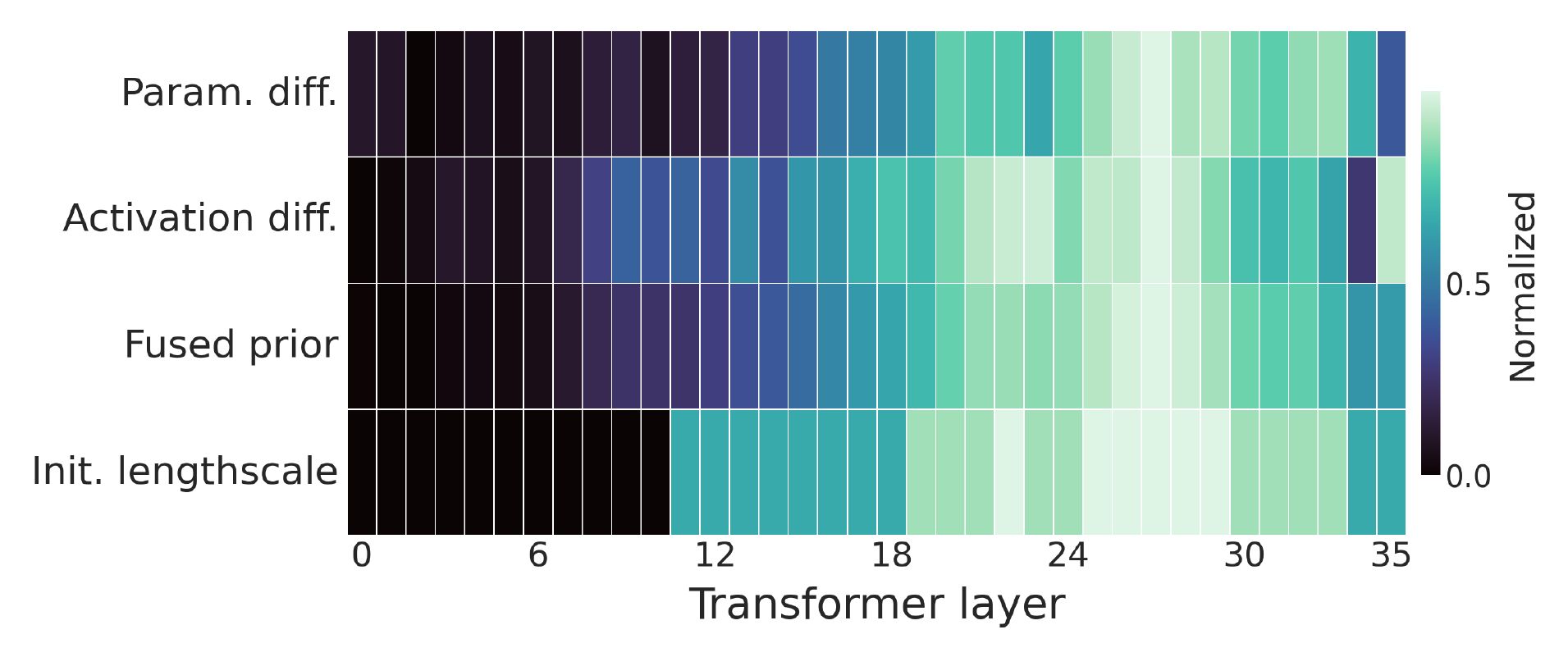}
		\caption{Layer-wise prior used in the main experiment. The heatmap reports parameter discrepancy, reasoning activation discrepancy, the smoothed fused score, and the resulting initial lengthscale assigned by the GP kernel; smaller lengthscales mean that the initial GP is more sensitive to that layer.}
		\label{fig:supp_prior_heatmap_main}
	\end{figure}

	Fig.~\ref{fig:supp_prior_heatmap_main} shows the layer-wise prior used in the main experiment. The first two rows report complementary evidence from the two task models: parameter-space discrepancy and reasoning-prompt activation discrepancy. Both signals are weak in the earliest layers and become stronger in the middle-to-late part of the Transformer, indicating that the layer-wise search should not be initialized as if every layer were equally informative. The fused prior smooths these two signals into a single layer-importance profile. The final row records the corresponding GP lengthscale initialization, which should be interpreted as a soft search bias: it changes where the surrogate is initially more sensitive, but it does not freeze any layer or remove any merge coefficient from the feasible space.

	\section{Representative Pooled Best Trade-off Solutions}
	\label{app:pooled_pareto_solutions}
	
	Table~\ref{tab:supp_exp1_individuals} reports representative pooled best trade-off solutions from the layer-wise merging comparison. These solution IDs are used by the qualitative examples in Section~\ref{app:qualitative_examples}. Higher $F_1$, $F_2$, and accuracies are better, while lower generated-token counts are better.

	\section{Qualitative Responses from Representative Evaluated Solutions}
	\label{app:qualitative_examples}
	
	This section provides qualitative responses from representative evaluated merged models selected to span high-capability, intermediate, balanced, and lower-cost regions of the capability--cost trade-off. These examples complement the quantitative results in the main paper by showing how different merged models answer the same benchmark question. The solution IDs refer to Table~\ref{tab:supp_exp1_individuals}.

	
	\onecolumn

	\begin{table}[htbp]
		\caption{Experiment 1: Metrics of Representative Pooled Best Trade-off Solutions.}
		\label{tab:supp_exp1_individuals}
		\begin{center}
			\resizebox{\textwidth}{!}{%
				\begin{footnotesize}
					\setlength{\tabcolsep}{3pt}
					\begin{tabular}{clcccccc}
							\toprule
							\textbf{Solution} & \textbf{Method} & \textbf{$F_1$} ($\uparrow$) & \textbf{$F_2$} ($\uparrow$) &
							\textbf{GPQA Acc.} ($\uparrow$) & \textbf{GPQA \# Tokens} ($\downarrow$) &
							\textbf{AIME25 Acc.} ($\uparrow$) & \textbf{AIME25 \# Tokens} ($\downarrow$) \\
							\midrule
							1 & AP-BMM (ours) & 0.9773 & 0.7300 & 0.6900 & 5295 & 0.7333 & 9704 \\
							2 & AP-BMM (ours) & 0.9046 & 0.7331 & 0.6800 & 5367 & 0.7000 & 9528 \\
							3 & AP-BMM (ours) & 0.6637 & 0.8256 & 0.6400 & 3497 & 0.6000 & 9987 \\
							4 & AP-BMM (ours) & 0.5773 & 0.8261 & 0.5800 & 3435 & 0.6333 & 10064 \\
							5 & AP-BMM (ours) & 0.2318 & 1.1506 & 0.4500 & 724 & 0.6000 & 6237 \\
							6 & AP-BMM (ours) & 0.2727 & 1.0674 & 0.4900 & 1503 & 0.5666 & 7100 \\
							7 & AP-BMM (ours) & 0.2454 & 1.0741 & 0.5000 & 805 & 0.5333 & 7926 \\
							8 & MOMM & 0.8000 & 0.7928 & 0.7000 & 5204 & 0.6000 & 8350 \\
							9 & MOMM & 0.8045 & 0.7881 & 0.6800 & 5022 & 0.6333 & 8719 \\
							10 & MOMM & 0.7546 & 0.8189 & 0.6800 & 4700 & 0.6000 & 8447 \\
							11 & MOEA/D + CMA-ES & 1.0319 & 0.5087 & 0.6700 & 8533 & 0.8000 & 10354 \\
							\bottomrule
					\end{tabular}
				\end{footnotesize}
			}
		\end{center}
	\end{table}
	
	\begin{longtable}{@{}>{\centering\arraybackslash}p{0.10\textwidth}>{\raggedright\arraybackslash}p{0.85\textwidth}@{}}
		\caption{Qualitative responses from representative evaluated merged models.\label{tab:supp_qualitative_responses}} \\
		\toprule
		\textbf{Solution ID} & \textbf{Content} \\
		\midrule
		\endfirsthead
		
		\multicolumn{2}{c}{\tablename\ \thetable\ -- \textit{Continued from previous page}} \\
		\toprule
		\textbf{Solution ID} & \textbf{Content} \\
		\midrule
		\endhead
		
		\midrule
		\multicolumn{2}{r}{\textit{Continued on next page}} \\
		\endfoot
		
		\bottomrule
		\endlastfoot
		
		\multicolumn{2}{p{0.95\textwidth}}{\textbf{Question (verbatim):} Answer the following multiple choice question. The last line of your response should be of the following format: 'ANSWER: \$LETTER' (without quotes) where LETTER is one of A,B,C,D. Think step by step before answering.

			You identified a new quorum-sensing peptide in the bacteria Lactobacillus acidophilus. You are highly interested if its role in intercellular communication is conserved among the eukaryotes. You treat yeast Saccharomyces cerevisiae with the peptide and interestingly observe the formation of the shmoo. You want to learn about the proteome of the active chromatin in the shmoo and recover it by chromatin immunoprecipitation followed by mass spectrometry. The least of proteins from which complex below will you observe in your assay?} \\
		\multicolumn{2}{p{0.95\textwidth}}{\textbf{Options:} A) nucleosome histone complex
			
			B) enhancer protein complex
			
			C) pre-replication complex
			
			D) pre-initiation complex} \\
		\multicolumn{2}{p{0.95\textwidth}}{\textbf{Correct Answer:} C} \\
		\midrule
		
		7 & 
		\textbf{Model:} Solution 7 in Table~\ref{tab:supp_exp1_individuals} (low-cost AP-BMM trade-off)
		
		\textbf{$F_1/F_2$:} 0.2454 / 1.0741
		
		\textbf{Generated tokens:} 321
		
		\textbf{Model Answer:} C
		
		\textbf{Generated Response:} Let's solve this step by step:

		1) The question is about identifying which protein complex is least likely to be observed in a chromatin immunoprecipitation (ChIP) assay of active chromatin in yeast shmoo formation.

		2) The shmoo structure in yeast is a polarized growth structure that forms in response to mating pheromones. This is a mating response, not a replication or transcription initiation event.

		3) Let's analyze each complex:
		
		- Nucleosome histone complex: This is a fundamental component of chromatin and present in all cells. It would be observed in any chromatin IP assay.
		
		- Enhancer protein complex: These are involved in gene regulation and are active in response to signals. In a mating response like shmoo formation, enhancers would be activated.
		
		- Pre-replication complex: This is involved in DNA replication initiation, which occurs during the cell cycle but not specifically in mating response.
		
		- Pre-initiation complex: This is involved in transcription initiation and would be active in response to mating signals. \\
		
		\midrule

		 &

		4) The key point is that the shmoo formation is a mating response, not a replication event. The pre-replication complex is specifically involved in DNA replication, which is not directly related to mating or shmoo formation.

		5) While all the other complexes might be involved in some way in the mating response, the pre-replication complex is least likely to be part of active chromatin in the shmoo structure because it's related to DNA replication, not transcription or mating signaling.

		ANSWER: C \\
		
		\midrule

		6 & 
		\textbf{Model:} Solution 6 in Table~\ref{tab:supp_exp1_individuals} (lower-cost AP-BMM trade-off)
		
		\textbf{$F_1/F_2$:} 0.2727 / 1.0674
		
		\textbf{Generated tokens:} 367
		
		\textbf{Model Answer:} C
		
		\textbf{Generated Response:} Let's solve this step by step:

		1) First, let's understand what we're looking for:
		
		- We're studying a quorum-sensing peptide in Lactobacillus acidophilus
		
		- The peptide induces shmoo formation in yeast (S. cerevisiae)
		
		- We want to study the proteome of active chromatin in the shmoo
		
		- We're using ChIP-seq followed by mass spectrometry

		2) Shmoo formation in yeast:
		
		- Shmoo formation is a well-known response to mating pheromones
		
		- It's a morphological change that occurs when yeast cells detect a mating partner
		
		- It involves activation of specific genes and chromatin changes

		3) What is active chromatin?
		
		- Active chromatin is chromatin that is accessible and transcriptionally active
		
		- It contains specific proteins that mark active regions

		4) Let's evaluate each complex:
		
		- A) Nucleosome histone complex: This is fundamental to all chromatin and present in all cells
		
		- B) Enhancer protein complex: These are involved in gene regulation and active chromatin
		
		- C) Pre-replication complex: This is involved in DNA replication, not active transcription
		
		- D) Pre-initiation complex: This is involved in transcription initiation

		5) Key point:
		
		- The pre-replication complex is involved in DNA replication, which occurs during S phase
		
		- Shmoo formation is a mating response, not a replication event
		
		- The pre-replication complex is not typically associated with active chromatin in transcriptionally active states

		6) Conclusion:
		
		- The pre-replication complex is least likely to be present in active chromatin during shmoo formation
		
		- The other complexes are more directly related to active chromatin and transcription

		ANSWER: C \\
		
		\midrule
		
				3 & 
		\textbf{Model:} Solution 3 in Table~\ref{tab:supp_exp1_individuals} (balanced AP-BMM trade-off)
		
		\textbf{$F_1/F_2$:} 0.6637 / 0.8256
		
		\textbf{Generated tokens:} 643
		
		\textbf{Model Answer:} C
		
		\textbf{Generated Response:} I need to think through this step by step.

		First, let me understand the context:
		
		- We have a quorum-sensing peptide from Lactobacillus acidophilus that causes shmoo formation in yeast (S. cerevisiae)
		
		- We're doing ChIP-seq to study the active chromatin in the shmoo
		
		- The question asks which complex we would least likely observe in this assay

		Key points about the assay:
		
		- It's chromatin immunoprecipitation (ChIP) followed by mass spectrometry
		
		- We're looking at active chromatin in the shmoo
		
		- This means we're specifically studying proteins that are bound to active chromatin regions

		Let me consider each option:

		A) Nucleosome histone complex - This is fundamental to all chromatin. All DNA is packaged in nucleosomes, so this complex would definitely be present in active chromatin.

		B) Enhancer protein complex - Enhancers are regulatory elements that control gene expression. In active chromatin, we would expect to see enhancer proteins bound to active regions.

		C) Pre-replication complex - This complex is involved in DNA replication initiation. It assembles at origins of replication during the G1 phase of the cell cycle. The shmoo formation is a morphological change in response to a signal, not directly related to DNA replication. In active chromatin, we're not typically looking at replication complexes.

		D) Pre-initiation complex - This complex assembles at transcription start sites to initiate transcription. In active chromatin, we would expect to see this complex bound to active genes.

		The key is that we're studying active chromatin in the context of a specific cellular response (shmoo formation). The pre-replication complex is involved in DNA replication, which is not directly related to the active chromatin state in a signaling response like shmoo formation. The other complexes (nucleosomes, enhancers, pre-initiation) are all directly involved in gene regulation and chromatin function.

		The pre-replication complex is least likely to be part of the active chromatin in this context because:
		
		1. It's specific to DNA replication
		
		2. The shmoo formation is a signaling response, not a replication event
		
		3. We're looking at active chromatin (which is about gene regulation, not replication)

		Therefore, the least likely complex to be observed in this active chromatin assay would be the pre-replication complex.

		ANSWER: C
		
		I need to double-check my reasoning. The question asks for the "least" of proteins from which complex we would observe. In active chromatin, we're looking at transcriptionally active regions. The pre-replication complex is involved in DNA replication, not in active transcription or chromatin states that are directly related to gene expression. The other complexes (nucleosomes, enhancers, pre-initiation) are all directly involved in chromatin function and gene regulation.

		Yes, the pre-replication complex is the one that's least likely to be part of the active chromatin in this context. The other options are all directly related to chromatin and gene regulation.

		ANSWER: C \\
		
		\midrule
		
		1 & 
		\textbf{Model:} Solution 1 in Table~\ref{tab:supp_exp1_individuals} (high-capability AP-BMM trade-off)
		
		\textbf{$F_1/F_2$:} 0.9773 / 0.7300
		
		\textbf{Generated tokens:} 1635
		
		\textbf{Model Answer:} C
		
		\textbf{Generated Response:} I need to figure out which protein complex is least likely to be observed in the chromatin immunoprecipitation (ChIP) assay for active chromatin in the shmoo of Saccharomyces cerevisiae after treatment with a quorum-sensing peptide from Lactobacillus acidophilus.

		First, recall that ChIP is used to identify proteins bound to specific DNA regions, like active chromatin. In this case, we're looking at the proteome of active chromatin in the shmoo, which is a yeast phenotype associated with mating.

		The shmoo formation in yeast is part of the mating process. When yeast cells are treated with this peptide, it induces shmoo formation, which is a morphological change for mating. So, the peptide is acting as a signal that triggers mating responses.

		Active chromatin in this context would refer to chromatin regions that are being transcribed during the mating response. So, ChIP would pull down proteins that are bound to active chromatin, like transcription factors, histones, and other chromatin-associated proteins.

		Now, let's look at the options:

		A) Nucleosome histone complex: This is a fundamental part of chromatin. Histones are always present in chromatin, and ChIP often involves histone antibodies to pull down nucleosomes. So, this complex should be very abundant in active chromatin. It's likely to be observed.

		B) Enhancer protein complex: Enhancers are regulatory elements that control gene expression. In active chromatin, enhancer proteins (like transcription factors binding to enhancers) are part of the active chromatin landscape. Since shmoo formation involves gene activation for mating, enhancers are probably involved. So, this complex should be observed.

		C) Pre-replication complex: This complex is involved in DNA replication initiation. It assembles at origins of replication during the cell cycle, before DNA replication. In yeast, the pre-replication complex (pre-RC) is active in the G1 phase, but it's not directly related to active transcription or mating responses. Shmoo formation is a mating response, not a replication event. The peptide treatment is inducing a mating signal, not replication. So, the pre-replication complex might not be part of active chromatin in this context. Active chromatin typically refers to transcriptionally active regions, not replication origins.

		D) Pre-initiation complex: This is the complex that assembles at promoters to initiate transcription. It includes RNA polymerase II and general transcription factors. In active chromatin, especially during transcriptional activation like in mating, the pre-initiation complex is crucial. Shmoo formation involves transcription of mating genes, so this complex should be present.

		The question asks for the "least" of proteins from which complex will be observed. So, I need to find which one is least likely to be in active chromatin during shmoo formation.

		- A: Histone complex - always in chromatin, so definitely observed.

		- B: Enhancer protein complex - enhancers are active in gene regulation for mating, so likely observed.

		- C: Pre-replication complex - this is for DNA replication, which isn't directly tied to the mating response. Shmoo formation is a signal for mating, not replication. In yeast, replication occurs in the cell cycle, but shmooing is a transient response. The peptide might not be triggering replication; it's a quorum-sensing peptide that induces mating. So, pre-replication complex is probably not part of active chromatin in this context.

		- D: Pre-initiation complex - directly involved in transcription initiation, which is key for mating gene expression. So, it should be observed.
		
		\\
		 & 
		
		I recall that in ChIP for active chromatin, we often see histones, transcription factors, and pre-initiation complexes. Pre-replication complexes are not typically associated with active transcription; they're for replication.

		In yeast, the pre-replication complex (pre-RC) includes proteins like Cdc6, Cdt1, MCM complex, which load onto origins. But active chromatin in the context of transcription refers to regions with open chromatin and transcription factors, not replication complexes.

		The shmoo is a mating structure, so it's about gene expression for mating, not replication. Therefore, the pre-replication complex shouldn't be part of the active chromatin proteome here.

		Let me double-check option B: enhancer protein complex. In yeast, enhancers are important for mating, like the mating pheromone response. So, enhancers are active during shmoo formation.

		Option D: pre-initiation complex is definitely involved in transcription.

		So, between C and the others, C seems the least relevant.

		The question says "the proteome of the active chromatin". Active chromatin in this context is for transcriptional activity, not replication.

		Therefore, the pre-replication complex is least likely to be observed.

		I think C is the answer.

		Just to be thorough, is there any chance that replication is happening during shmoo? Shmoo formation is a morphological change, not a replication event. Yeast cells might be in G1 phase when mating, so replication isn't happening. The peptide is inducing a signaling pathway for mating, not replication.

		So, yes, pre-replication complex should not be part of this active chromatin assay.

		The other complexes are directly involved in transcription or chromatin regulation.

		Therefore, the least observed would be the pre-replication complex.

		ANSWER: C
		
		To determine which protein complex is least likely to be observed in the chromatin immunoprecipitation (ChIP) assay for active chromatin in the shmoo of *Saccharomyces cerevisiae* after treatment with the quorum-sensing peptide, consider the context:

		- The peptide induces shmoo formation, a mating-specific response in yeast. This involves transcriptional activation of mating genes (e.g., pheromone response genes).
		
		- Active chromatin in this context refers to regions dynamically regulated during transcription, such as promoters and enhancers, where transcription factors and RNA polymerase complexes are bound.
		
		- ChIP for active chromatin targets proteins associated with transcriptionally active regions.

		Now evaluate each option:

		- **A) Nucleosome histone complex**: Histones are fundamental components of chromatin and are always present in active chromatin. ChIP assays often enrich for histones (e.g., H3K4me3 for active promoters). This complex will be abundantly observed.
		
		- **B) Enhancer protein complex**: Enhancers regulate gene expression during mating responses. In yeast, enhancers (e.g., for mating genes) are activated during shmoo formation, so transcription factors binding to enhancers are part of active chromatin. This complex is expected to be observed.
		
		- **C) Pre-replication complex**: This complex (e.g., Cdc6, Cdt1, MCM) assembles at DNA replication origins during the G1 phase to initiate replication. It is not involved in transcriptional activation or mating responses. Shmoo formation is a mating signal (transcriptional response), not a replication event. In yeast, mating occurs without immediate replication, and active chromatin in this context specifically refers to transcriptionally active regions, not replication origins. Thus, this complex is least likely to be part of the active chromatin proteome in this assay.
		
		- **D) Pre-initiation complex**: This complex (e.g., RNA polymerase II, TFIID, TFIIB) assembles at promoters to initiate transcription. Mating gene expression requires this complex, so it is directly involved in active chromatin during shmoo formation and will be observed.

		The question asks for the **least** observed complex. Given that the shmoo response is transcriptional and mating-specific, the pre-replication complex is unrelated to active transcriptional chromatin and thus would not be enriched in this ChIP assay.

		ANSWER: C \\
		
		\midrule
		
	\end{longtable}